\documentclass[conference]{IEEEtran}
\IEEEoverridecommandlockouts
\usepackage{cite}
\usepackage{amsmath,amssymb,amsfonts}
\usepackage{algorithmic}
\usepackage{graphicx}
\usepackage{textcomp}
\usepackage{xcolor}

\usepackage{hyperref}
\usepackage{algorithm}
\usepackage{framed}
\usepackage{caption}
\usepackage{subcaption}
\usepackage{booktabs}
\usepackage{array}
\usepackage{multirow}

\def\BibTeX{{\rm B\kern-.05em{\sc i\kern-.025em b}\kern-.08em
    T\kern-.1667em\lower.7ex\hbox{E}\kern-.125emX}}

\begin{document}

%iqrah
\makeatletter
\newcommand{\linebreakand}{%
  \end{@IEEEauthorhalign}
  \hfill\mbox{}\par
  \mbox{}\hfill\begin{@IEEEauthorhalign}
}
\makeatother

\title{A Parallel Workflow for Polar Sea-Ice Classification using Auto-labeling of Sentinel-2 Imagery
%Parallel Workflow for Polar Sea-Ice Classification using Auto-labeling of Sentinel-2 Imagery to Train an Efficient Deep Learning Model
}

% \and
% \and
% \and
% % \and
% % \IEEEauthorblockN{5\textsuperscript{th} Given Name Surname}
% % \IEEEauthorblockA{\textit{dept. name of organization (of Aff.)} \\
% % \textit{name of organization (of Aff.)}\\
% % City, Country \\
% % email address or ORCID}
% % \and
% % \IEEEauthorblockN{6\textsuperscript{th} Given Name Surname}
% % \IEEEauthorblockA{\textit{dept. name of organization (of Aff.)} \\
% % \textit{name of organization (of Aff.)}\\
% % City, Country \\
% % email address or ORCID}
% }

\author{
   \IEEEauthorblockN{
   %1\textsuperscript{st}
   Jurdana Masuma Iqrah}
   \IEEEauthorblockA{\textit{Department of Computer Science} \\
       \textit{University of Texas at San Antonio}\\
       San Antonio, Texas, USA \\
       %email address or ORCID
   }
   % \and
   % \IEEEauthorblockN{
   % %3\textsuperscript{rd} 
   % Younghyun Koo}
   % \IEEEauthorblockA{\textit{Department of Earth and Planetary Sciences} \\
   %     \textit{University of Texas at San Antonio}\\
   %     San Antonio, Texas, USA \\
   %     email address or ORCID
   % }
   \and
   \IEEEauthorblockN{
   %2\textsuperscript{nd} 
   Wei Wang}
   \IEEEauthorblockA{\textit{Department of Computer Science} \\
       \textit{University of Texas at San Antonio}\\
       San Antonio, Texas, USA \\
       %email address or ORCID
   }
   %\and
   %iqrah
   \linebreakand % <------------- \and with a line-break 
   \IEEEauthorblockN{
   %4\textsuperscript{th} 
   Hongjie Xie}
   \IEEEauthorblockA{\textit{Department of Earth and Planetary Sciences} \\
       \textit{University of Texas at San Antonio}\\
       San Antonio, Texas, USA \\
       %email address or ORCID
   }
   \and
   \IEEEauthorblockN{
   %5\textsuperscript{th} 
   Sushil Prasad}
   \IEEEauthorblockA{\textit{Department of Computer Science} \\
       \textit{University of Texas at San Antonio}\\
       San Antonio, Texas, USA \\
       %email address or ORCID
   }
}

\maketitle

\begin{abstract}
%Global warming is an urgent issue generating catastrophic environmental impacts, such as melting sea ice and glaciers, particularly in polar regions. The observation of the advancing and retreating pattern of polar sea ice cover stands as a vital indicator of the ongoing impact of global warming.

%Global warming is an urgent issue generating catastrophic environmental impacts, particularly in polar regions. 
The observation of the advancing and retreating pattern of polar sea ice cover stands as a vital indicator of global warming.
This research aims to develop a robust, effective, and scalable system for classifying polar sea ice as thick/snow-covered, young/thin, or open water using Sentinel-2 (S2) images.
Since the S2 satellite is actively capturing high-resolution imagery over the earth's surface, there are lots of images that need to be classified.
One major obstacle is the absence of labeled S2 training data (images) to act as the ground truth.
We demonstrate a scalable and accurate method for segmenting and automatically labeling S2 images using carefully determined color thresholds.
We employ a parallel workflow using PySpark to scale and achieve 9-fold data loading and 16-fold map-reduce speedup on auto-labeling S2 images based on thin cloud and shadow-filtered color-based segmentation to generate label data.
The auto-labeled data generated from this process are then employed to train a U-Net machine learning model, resulting in good classification accuracy.
As training the U-Net classification model is computationally heavy and time-consuming, we distribute the U-Net model training to scale it over 8 GPUs using the Horovod framework over a DGX cluster with a 7.21x speedup without affecting the accuracy of the model.
Using the Antarctic's Ross Sea region as an example, the U-Net model trained on auto-labeled data achieves a classification accuracy of 98.97\% for auto-labeled training datasets when the thin clouds and shadows from the S2 images are filtered out.

\end{abstract}

\begin{IEEEkeywords}
Polar Sea Ice, Sentinel-2, Sea Ice Classification, Auto-labeling, Parallel Processing, Distributed Deep Learning, Synchronous Data Parallel
\end{IEEEkeywords}

\section{Introduction}\label{sec:introduction}
%\section{Introduction}\label{sec:introduction}
% for  global warming sea ice study needed
Global warming is an urgent issue generating catastrophic environmental impacts, particularly in polar regions. Eventually, it will lead to a rise in sea level, resulting in coastal land loss, a shift in precipitation patterns, increased drought and flood risks, and threats to biodiversity {\cite{serreze2011processes}}. Therefore, assessing the melting and retreat of polar sea ice cover is key as an indicator of global warming. 

%lots of image
Since the S2 satellite is actively capturing high-resolution imagery over the earth's surface, there are lots of images that need to be classified. This big data processing requires high computation power, high performance, and scalability.

%absence of annotated data
Again an early challenge to using S2 images to classify sea ice is the lack of annotated/labeled polar sea ice cover data for training and validation. Because sea ice typically has irregular shapes, and also because of the existence of clouds and shadows in the images, it usually requires careful manual labeling to mark different types of sea ice in the S2 images. This manual labeling is extremely time-consuming and thus not scalable. We observed that the manual labeling by Earth scientists is typically based on the color of the image pixels (e.g., large white areas are usually thick/snow-covered ice). 

This inspired us to explore a scalable color-based segmentation to automatically label S2 polar sea ice images. By observing the collected S2 dataset, we found that the color ranges for polar sea ice and open water are almost constant for the summer season.
This project presents our preliminary results on a parallel workflow for color-based segmentation, auto-labeling, and polar sea ice classification
\footnote{A preliminary version of this manuscript has been presented as a non-archival poster in the \href{https://ai-2-ase.github.io/papers/}{AI2ASE workshop} of AAAI 2023\cite{iqrahtoward}}.

We acquired 4224 images of 256x256 pixel images from the Sentinel-2 dataset from the Ross Sea region of Antarctica during the polar summer season for training machine learning models. Our technique initially removes thin clouds and shadows from the images. After that, our color-segmentation algorithm labels sea ice cover images automatically. This stage creates three masks representing snow-covered or thick ice, thin ice, and open water based on color ranges. The color-segmentation method automatically annotates polar sea ice and open water or leads for training a U-net model.
To scale the auto-labeling process using shadow-cloud-filtered color-based segmentation, we initially applied the Python multiprocessing library yielding 4.5x speedup. We then utilized a PySpark-based map-reduce to further scale the thin cloud and shadow-removal and auto-labeling processes. This {\it Map-Reduce}-based scaling increased the speedup to 16.25x for four executor nodes with four cores each employing a Google Cloud Dataproc (GCD) cluster with a master node and three worker nodes configuration.
We tested the effectiveness of auto-labeling by training two U-net models first using the manually labeled data ({\bf U-Net-Man}) and second using the auto-labeled data ({\bf U-Net-Auto}), then validating these two models on the same manually-labeled dataset. 
We scaled the U-net model training using the Horovod framework. This distributed model training achieved a 7.21x speedup on an 8-GPU DGX cluster. 
Our U-Net-Auto model achieves  98.97\% accuracy over original S2 images after thin cloud and shadow filtered out. For the U-Net-Man model, the corresponding accuracy is 98.40\%.

%\textcolor{red}{why we used different types of scaling techniques for different portions}

The key contributions of this work are
\begin{itemize}
    \item a parallelized S2 imagery auto-labeling gaining 4.5x speedup with Python Multiprocessing and 16.25x speedup employing PySpark; 
    \item a Horovod-based distributed deep learning model training with an almost linear speedup of 7.21x on 8 GPUs, and
    \item a U-Net model trained on the auto-labeled data yielding 98.97\% accuracy.
\end{itemize}

%We plan to expand the scope of this research to include more regions and seasons in future research projects. This extension will contribute to strengthening the validity of our hypotheses on the auto-labeling of the corresponding training datasets. It is, therefore, expected that our U-net model can be trained further to demonstrate robustness against temporal and spatial variations. Finally, we will employ the scaling methodologies to increase speed.

%\footnote{

%To reproduce our experimental results we made our source code and sample datasets available on GitHub. The link is provided in reference \cite{IqrahSentinel-2Sea-IceClassification2022}.
%}

%Paper organization
The remaining sections of this paper are organized in the following order. Section 2 reviews the key related work. Section 3 describes our methodology for the parallel sea ice auto-labeling, model training, and classification workflow. Section 4 contains the evaluation metrics, experimental results, and a discussion of the proposed methodologies. In Section 5, we provide concluding remarks and suggest future directions for this ongoing work.

To reproduce our experimental results we made our
source code and sample datasets available on GitHub
\footnote{
\href{https://github.com/jmiqra/S2_Parallel_Workflow}{S2 parallel workflow github code  link}
}.

\section{Related Work}\label{sec:rel_work}
%\section{Related Work}\label{sec:rel_work}
Researchers have observed the sea ice melting and retreat, particularly in the arctic, and projected that this amplification is expected to get stronger over time \cite{serreze2011processes}.
Earlier, researchers relied on space-borne satellite data like Sentinel-1 ({\it S1}) Synthetic Aperture Radar (SAR) for extracting sea ice information  \cite{sentinel1sar}. Afterward, Sentinel-2 ({\it S2}) was launched in 2015, offering higher-resolution optical images than S1 SAR images. S2 optical images - that we have employed - have up to 10m high spatial resolution with finer-grain and more detailed sea-ice images compared to S1 with 40m spatial resolution. 
In \cite{park2020classification}, authors proposed Haralick texture features and random forests classifier to retrieve several ice types. However, with around 85\% accuracy, this model is computationally complex and is sensitive to texture noise \cite{8630667} and thermal noise \cite{8126233} in Sentinel-1 data. \cite{boulze2020classification} applied a sea ice types classification using a Convolutional Neural Network (CNN) on Sentinel-1 dual Horizontal-Horizontal (HH) and Horizontal-Vertical (HV) polarization. They successfully trained their CNN and could retrieve four classes of ice: old ice, first-year ice, young ice, and ice-free (open water) with around 90\% accuracy. For ice types' names and codes, they followed the World Meteorological Organization (WMO)  codes \cite{joint2014ice}. They also compared their results with an existing random forest classification \cite{park2020classification} for each ice type and proved theirs to be more efficient based on execution time and less noise sensitivity on SAR data. 
Furthermore, to derive a high-resolution sea ice cover product for the Arctic using S1 dual HH and HV polarization data in extra wide swath (EW) mode, a U-net model was applied in \cite{wang2021arctic}. They used S1 images with 40m spatial resolution to classify sea ice and open water.

\cite{campos2020understanding} used S2 time series to classify land use using a 2-BiLSTM recurrent neural network model. 
\cite{muchow2021lead} introduced the application of a sea-ice surface type classification on 20 carefully selected cloud-free S2 Level-1C products. To detect {\it leads} (narrow, linear cracks in the ice sheet), they first created five sea-ice surface type classifications (open water, thin ice - nilas, gray sea ice, gray-white sea ice, and sea ice covered with snow); these names are based on the WMO \cite{joint2014ice} for uniformity with other literature. However, they manually masked each five surface classes to get the  Top of Atmosphere (TOA) reflectance value dataset for each mask. 
Another sea-ice monitoring \cite{wang2021monitoring} work on S2 data, in Liaodong Bay, Bohai sea with less than 10\% cloud cover, trained their decision tree on different sea-ice types from Normalized Difference Snow Index (NDSI) and the Normalized Difference Vegetation Index (NDVI). This region has seasonal sea ice with less thickness than the sea ice in the polar regions.

In \cite{hernandez2020exploring} authors have utilized multi-core parallel processing for land cover classification using a random forest classifier model and achieved 76\% accuracy. Again, there is FORCE \cite{frantz2019force}, an open software developed in C/C++ that parallelizes each data cube of Landsat 7/8 and S2 A/B using within the tile multithreading. This paper \cite{zhang2020super} proposed a distributed deep learning model for unknown low to high-resolution mapping of large-volume S2 data. They utilized synchronous data-parallel \cite{wiki:dataparallel} distributed training via Horovod over high-performance computing systems with GPUs.

%Parallel processing is the practice of carrying out multiple tasks at once within a predetermined window of time. Its goal is to use several processors to reduce computing time. Using the pool object, the multiprocessing package in Python programming supports parallelizing the execution of functions with multiple input values spread across processes.
%MapReduce is a programming model and processing technique designed to handle large-scale data processing tasks in a parallel and distributed manner.

\section{Methodology}\label{sec:methodology}
%\section{Methodology}\label{sec:methodology}

%\subsection{Methodology Overview}
%We focus on the polar sea ice regions to study the sea ice retreat and melting. Hence, we seek to identify and classify different sea ice types and open water using the S2 satellite imagery from the polar region.

This study on parallel workflows involves the parallelization and distribution of two key components: auto-labeling and U-Net model training for the classification of polar sea ice. Figure \ref{fig:para_workflow} depicts the parallel and distributed auto-labeling and U-Net model training workflow along with the underlying architectures.

\begin{figure}[ht]
    \begin{framed}
        \centering
        \includegraphics[width=\textwidth]{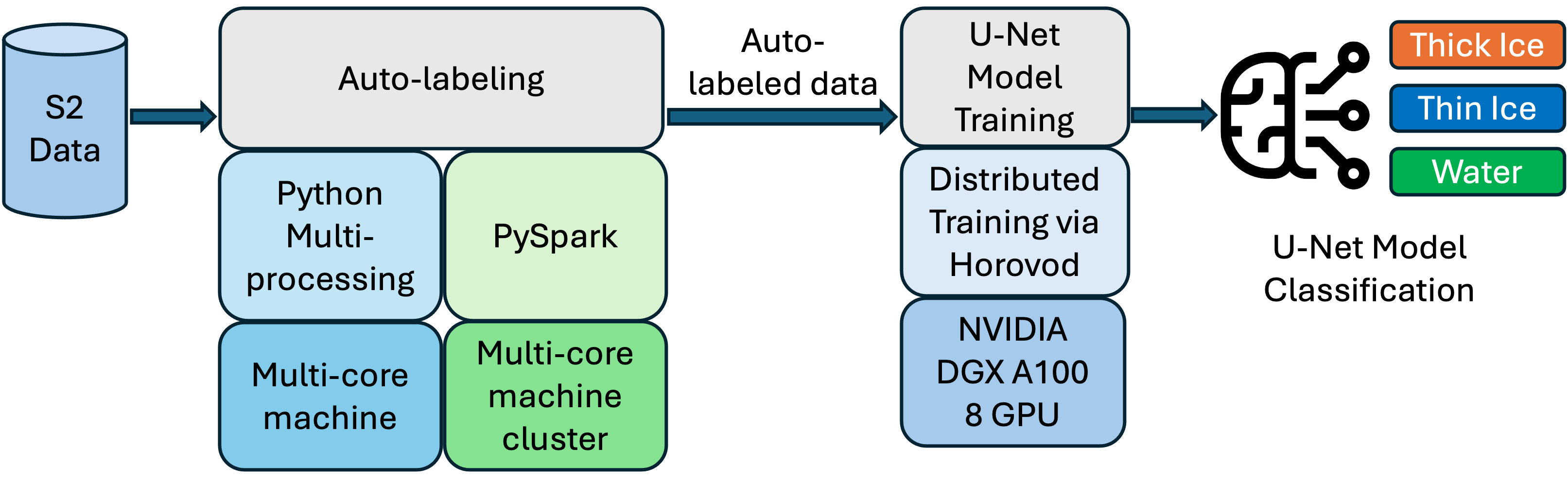}
    \end{framed}
    \caption{Workflow for parallel and distributed auto-labeling and U-Net model training.}
    \label{fig:para_workflow}
\end{figure}
%\textcolor{red}{\tt Figure 1 font size needs increasing - if needed, full two column may be used.  Same problem with figure 5.}

There are three main types of sea ice for classification: thick ice, thin ice, and open water. Identifying these types of sea ice cover will help observe significant changes in the polar sea ice. S2 satellite captures high-resolution optical imagery over land and coastal waters, including the polar sea ice regions. These are more fine-grained images with 10m to 60m spatial resolution. However, these images often include clouds and shadows of clouds that affect the clarity of the corresponding pixels of the image. As a result, the clouds/shadows affect the sea ice segmentation and classification. Therefore, to address this problem, we apply image transformation techniques to remove shadows and thin clouds from the cloudy and shadowy images and store them as thin {cloud and shadow-filtered images}. 

%\textcolor{red}{write more according to the work flow, change workflow}
%The figure \ref{fig:workflow} includes the overall workflow of the proposed methodology.

\begin{figure}[ht]
    \begin{framed}
        \centering
        \includegraphics[width=\textwidth]{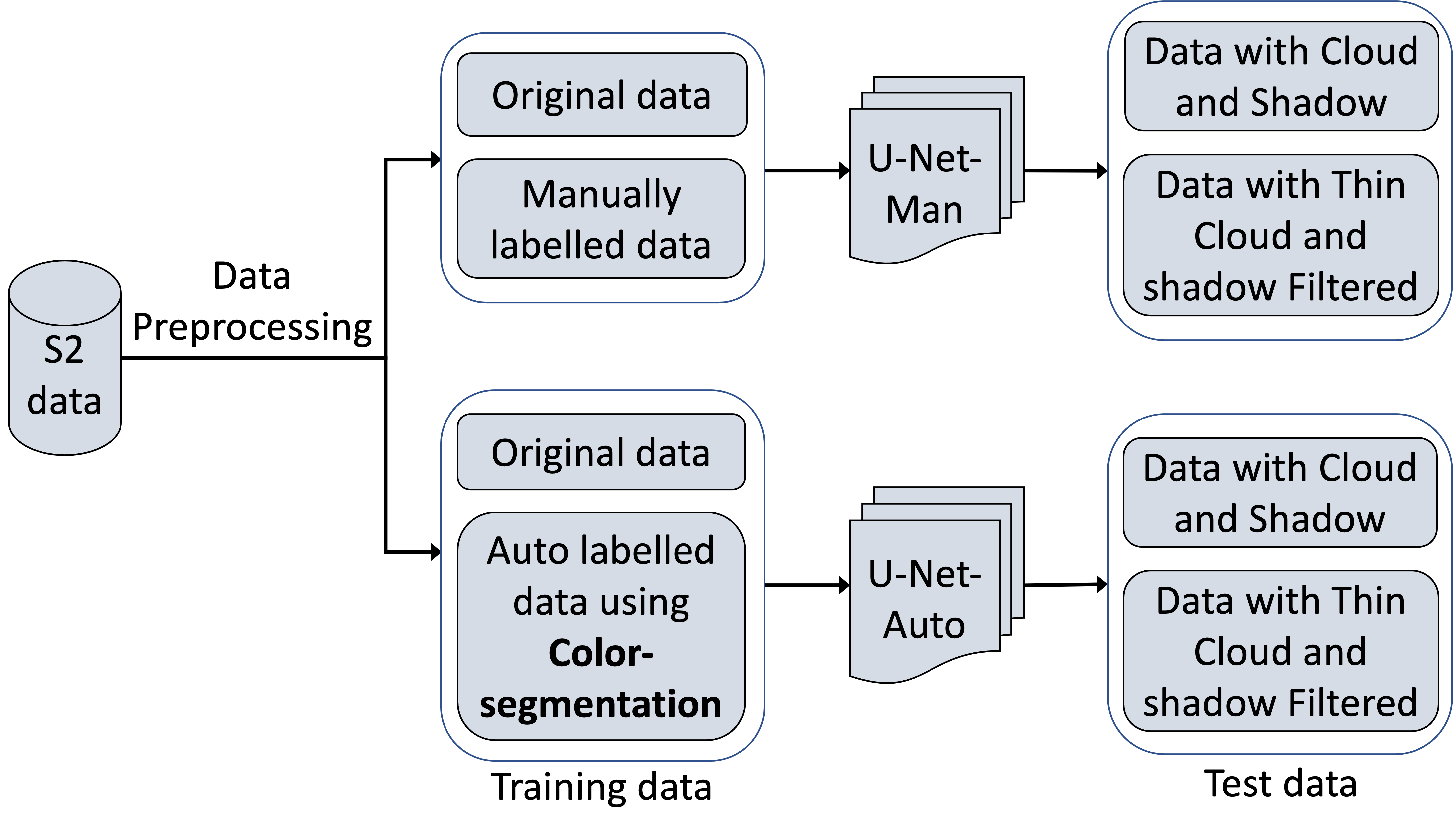}
    \end{framed}
    \caption{Workflow for training and test data preparation and sea ice classification methodology.}
    \label{fig:workflow}
\end{figure}

A key problem is that there is no labeled sea ice cover data available for training a model. To address this, we employ the workflow as shown in Figure \ref{fig:workflow}. After collecting the S2 imagery for the Ross sea region in the Antarctic, we first manually label the data, mainly for validation purposes. 
Then, we also apply color-based segmentation to label different sea ice and open water based on their pixels' color threshold limit values, which yields auto-labeled S2 sea-ice cover data.
%As long as the colors for the sea ice and open water in the current set of the polar images fall in common ranges, the color-based segmentation will work well for this segmentation problem.

%\textcolor{red}{\tt You may add a third row for color coding in Figure 3, visually showing the three regions, either as arrows pointing to the regions in the second row or legends of three colors labeled as thick ice, thin ice, and open water}
Then, we train two U-Net models with deep neural network-based architecture, one on manually labeled images (U-Net-Man) and the other on auto-labeled images (U-Net-Auto). Then, we compare their accuracy results based on both (i) the original S2 data, including those with cloud and shadows, and (ii) cloud and shadow filtered data to validate our auto-labeling process.  We now describe these in detail.

\subsection{Sentinel-2 Data Collection and Preprocessing}
%Sentinel-2 is an Earth observation mission from the Copernicus Program by European Spatial Agency (ESA). It effectively captures optical imagery of land and coastal waters with high spatial resolution. 
In our workflow, we first select a spatial and temporal extent.  For our experiments, we choose the well-known spatial region in the Antarctic pole (Ross Sea), with spatial extent latitude (south) -70.00 to -78.00 and longitude (west) -140.00 to -180.00. For the temporal extent, we chose November 2019 data, which is the summer season.
Then, we collect the S2 satellite imagery for that spatial region and specific time using Google Earth Engine (GEE). Each image has 13 available bands; among those available bands, we select bands 2, 3, and 4, representing blue, green, and red, respectively. Each of these bands has a resolution of 10m.

Figure \ref{fig:s2 and ross sea} demonstrates two sample scenes of S2, one with cloud/shadow and one without cloud/shadow. We collected 66 large scenes and split them into over 4,224 images with 256x256 pixels each.
\begin{figure}[ht]
    \begin{framed}
        \centering
        %\begin{subfigure}[b]{0.2\textwidth}
        %    \centering
        %    \includegraphics[width=\textwidth]{figures/sentinel-2 satellite.jpeg}
        %    \caption{}
        %    \label{fig:s2}
        %\end{subfigure}
        %\hfill
        %\begin{subfigure}[b]{0.2\textwidth}
        %    \centering
        %    \includegraphics[width=\textwidth]{figures/ross_sea.jpg}
        %    \caption{}
        %    \label{fig:ross sea}
        %\end{subfigure}
        %\hfill
        \begin{subfigure}[b]{0.45\textwidth}
            \centering
            \includegraphics[width=\textwidth]{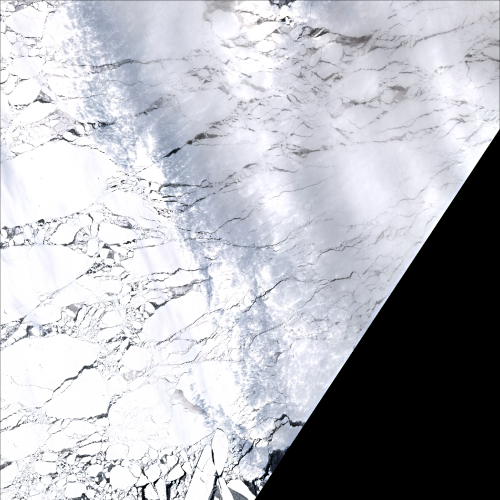}
            \caption{}
            \label{fig:s2_1}
        \end{subfigure}
        \hfill
        \begin{subfigure}[b]{0.45\textwidth}
            \centering
            \includegraphics[width=\textwidth]{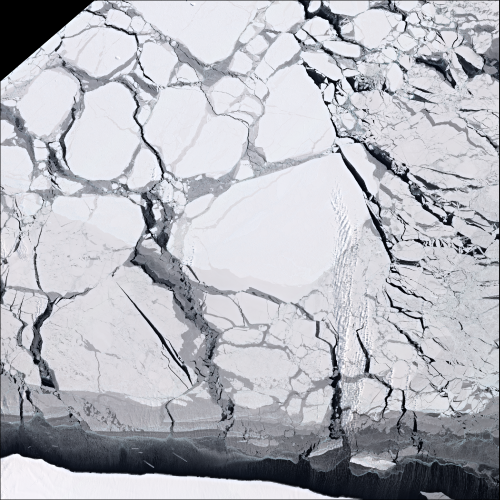}
            \caption{}
            \label{fig:s2_2}
        \end{subfigure}
    \end{framed}
    \caption{Sample Sentinel-2 scenes, (a) with thin cloud/shadow cover, and (b) without cloud or shadows.
    %a Sentinel-2 Satellite \cite{WikipediaEN:S2}, b Ross Sea region \cite{antarcticatravelcentre:rs} in the antarctic map with c and d Sample sentinel-2 scenes with and without cloud cover.
    }
    \label{fig:s2 and ross sea}
\end{figure}
Furthermore, we manually label our Sentinel-2 dataset to test and validate our methodology. We use red for snow-covered/thick ice, blue for thin or young ice, and green for open water regions, as illustrated in Figure \ref{fig:ori and manual}. 

\begin{figure}[ht]
    \begin{framed}
        \centering
        \begin{subfigure}[b]{0.3\textwidth}
            \centering
            \frame{\includegraphics[width=\textwidth]{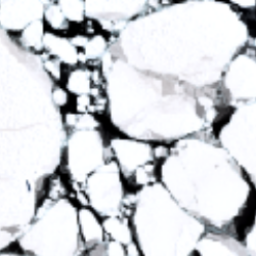}}
            \caption{}
            \label{fig:ori1}
        \end{subfigure}
         \hfill
        \begin{subfigure}[b]{0.3\textwidth}
            \centering
            \frame{\includegraphics[width=\textwidth]{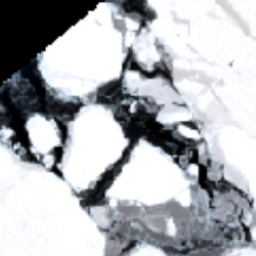}}
            \caption{}
            \label{fig:ori2}
        \end{subfigure}
         \hfill
        \begin{subfigure}[b]{0.3\textwidth}
            \centering
            \frame{\includegraphics[width=\textwidth]{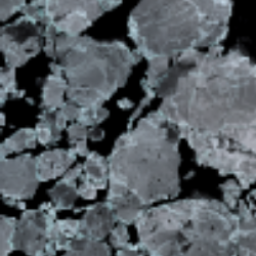}}
            \caption{}
            \label{fig:ori3}
        \end{subfigure}
        %\par\medskip        
        \begin{subfigure}[b]{0.3\textwidth}
            \centering
            \frame{\includegraphics[width=\textwidth]{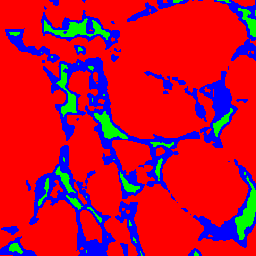}}
            \caption{}
            \label{fig:manu1}
        \end{subfigure}
        \hfill
        \begin{subfigure}[b]{0.3\textwidth}
            \centering
            \frame{\includegraphics[width=\textwidth]{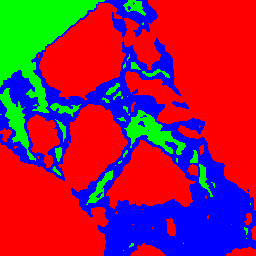}}
            \caption{}
            \label{fig:manu2}
        \end{subfigure}
         \hfill
        \begin{subfigure}[b]{0.3\textwidth}
            \centering
            \frame{\includegraphics[width=\textwidth]{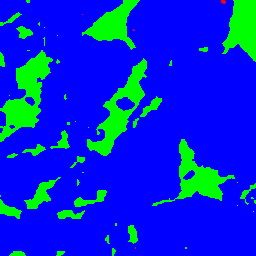}}
            \caption{}
            \label{fig:manu3}
        \end{subfigure}
        %\par\medskip
        \begin{subfigure}[b]{0.9\textwidth}
            \centering
            \includegraphics[width=\textwidth]{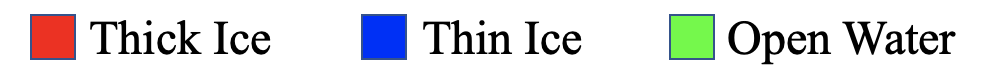}
        \end{subfigure}
        
    \end{framed}
    \caption{Manually-labeled data and their color codes, (a), (b), and (c) represent original Sentinel-2 data, (d), (e), and (f) represent the respective manually-labeled data. 
    %Here, in the manually labeled data, red, blue, and green colors, respectively, represent thick ice, thin ice, and open water.
    }
    \label{fig:ori and manual}
\end{figure}

\begin{figure}[htb]
    \begin{framed}
        \centering
        \begin{subfigure}[b]{0.3\linewidth}
            \centering
            \frame{\includegraphics[width=\linewidth]{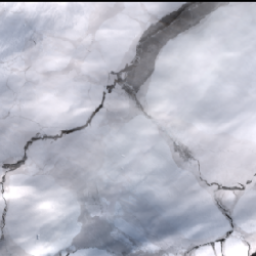}}
            \caption{}
            \label{fig:shdw12}
        \end{subfigure}
        \hfill
        \begin{subfigure}[b]{0.3\linewidth}
            \centering
            \frame{\includegraphics[width=\linewidth]{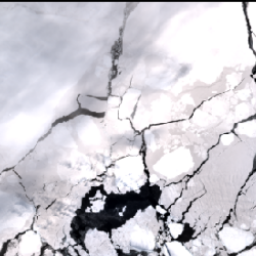}}
            \caption{}
            \label{fig:shdw22}
        \end{subfigure}
        \hfill
        \begin{subfigure}[b]{0.3\linewidth}
            \centering
            \frame{\includegraphics[width=\linewidth]{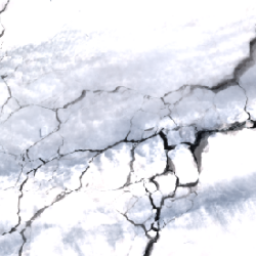}}
            \caption{}
            \label{fig:shdw32}
        \end{subfigure}
        %\par\medskip
        \begin{subfigure}[b]{0.3\linewidth}
            \centering
            \frame{\includegraphics[width=\linewidth]{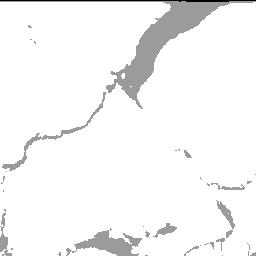}}
            \caption{}
            \label{fig:ref12}
        \end{subfigure}
        \hfill
        \begin{subfigure}[b]{0.3\linewidth}
            \centering
            \frame{\includegraphics[width=\linewidth]{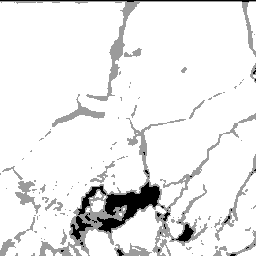}}
            \caption{}
            \label{fig:ref22}
        \end{subfigure}
        \hfill
        \begin{subfigure}[b]{0.3\linewidth}
            \centering
            \frame{\includegraphics[width=\linewidth]{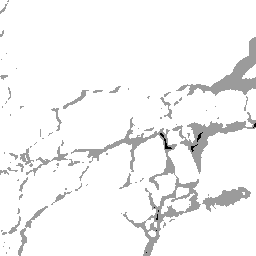}}
            \caption{}
            \label{fig:ref32}
        \end{subfigure}   
    \end{framed}
    \caption{Thin Cloud and Shadow Filtered Dataset, (a), (b) and (c) represent Sentinel-2 thin cloudy and shadowy images and (d), (e) and (f) as the corresponding filtered images.}
    \label{fig:shadow and cloud2}
\end{figure}

\subsection*{Filtering Out the Thin Clouds and Shadows}
The quality of S2 images is frequently compromised by the presence of both dense and sparse shadows, as well as cloud cover. Consequently, the presence of cloud cover impedes the ability to visually observe the sea ice cover on the surface of the polar zone. In situations where shadows and clouds exhibit high density, the complete elimination of these elements from a picture becomes challenging in the absence of equivalent reference shadows and cloud-free ground truth images of the identical location, derived from either the same or separate satellite image sources \cite{sarukkai2020cloud, meraner2020cloud}. One issue pertains to the fact that S2 revisits the same geographic area every five days, but sea ice coverage has the potential to undergo alterations during this time frame.

Nevertheless, the terrestrial surface is partially discernible by the presence of faint shadows and atmospheric formations. The utilization of radiation and brightness parameters has been found to be effective in the removal of thin clouds from an image \cite{liu2014thin}. Therefore, we decided to apply image transformations to filter our shadowy and cloudy images. For image transformation, we employ a range of known techniques available in OpenCV library \cite{opencv_library}, including RGB to HSV format conversion, noise filtering, bit-wise operations, absolute difference, Otsu thresholding \cite{xu2011characteristic}, Truncated thresholding \cite{guruprasad2020overview} and Binary thresholding and min-max normalization.  
Figure \ref{fig:shadow and cloud2}, displays images that are characterized by their thin, cloudy, and shadowy appearance, alongside the equivalent filtered images. The utilization of a cloud-and-shadow filtered dataset enhances the accuracy of sea ice and open water labels and classification in both color-segmentation-based labeling and U-net model testing.

For the auto-labeling of S2 images, we use color-based image segmentation. The figure \ref{fig:col_seg_archi} demonstrates the auto-labeling workflow in detail.

\subsection{Auto-labeling Method}
\begin{figure*}[htb]
    %\begin{framed}
        \centering
        \frame{\includegraphics[width=\linewidth]{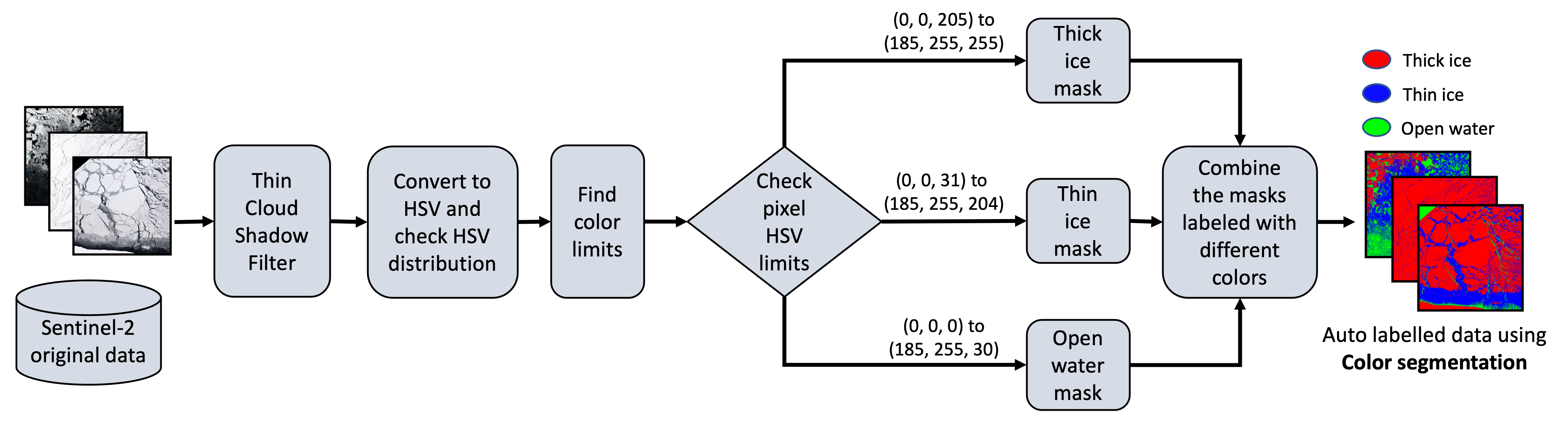}}
    %\end{framed}
    \caption{Auto-labeling Workflow: To label thick ice, thin ice, and open water from Sentinel-2 optical imagery for training data preparation.}
    \label{fig:col_seg_archi}
\end{figure*}
%\textcolor{red}{\tt Figure 5 comments: 3rd box: Find color...; 4th box: Check pixel HSV limits; 3rd branch: Open water [no dash]; last box: ...colors\\
%Also, fonts need to be larger, consider full two column} 
%\subsection{Color-based Image Segmentation}
Initially, the thin cloud and shadow present in the original image are subjected to a filtering process. 
Upon converting the imagery to the HSV format, the HSV value distribution of the sea ice segment from the current data set is utilized to manually ascertain a general color range for each sea ice type through a process of trial and error. This involves determining both a lower limit and an upper limit for the color range. 
In the context of our Antarctic Ross Sea imagery during the summer season, the HSV lower and upper values for thick ice range from (0, 0, 205) to (185, 255, 255). Similarly, for thin ice, the HSV lower and upper values span from (0, 0, 31) to (185, 255, 204). Lastly, the HSV lower and upper values for open water are defined as (0, 0, 0) to (185, 255, 30). 
The borders under consideration exhibit non-intersecting characteristics and can be readily evaluated against individual pixels. Using the established limits of the HSV color space, we generate distinct masks for each class.
In order to tackle the issue of picture labeling and annotation, a solution is proposed wherein the masks, each assigned a distinct color, are merged together. This merged dataset is then utilized for color-based segmentation on the S2 sea ice dataset. 
In this context, three distinct colors are assigned to represent three distinct categories: thick ice, thin ice, and open water segments, respectively. 
Color segmentation is employed in order to acquire the auto-labeled S2 sea-ice cover data pertaining to the Ross Sea region situated in the Antarctic.

\subsection*{Scaling Auto-labeling}
To scale the auto-labeling process comprising thin cloud and shadow filtering followed by color-based segmentation, we explored Python Multiprocessing for a single machine and PySpark Map-Reduce-based distributed technique for multiple heterogeneous machines.

%\subsubsubsection{Python Multiprocessing}
\paragraph{Python Multiprocessing}
%Parallel processing is the practice of carrying out multiple tasks at once within a predetermined window of time. Its goal is to use several processors to reduce computing time. 
In order to enhance the efficiency of the auto-labeling process over a reasonably large dataset consisting of 4224 images, each with 256x256 pixels, we started with a simple but effective multi-core-based parallelization on a workstation, which resulted in improved speedup.
%The auto-labeling segmentation approach employed in this study demonstrates a high degree of scalability. 
%To enhance the performance of the process on a desktop, Python multiprocessing module was utilized to parallelize and scale the operations, which resulted in improved speedup. 
%Using the pool object, the multiprocessing package in Python programming supports parallelizing the execution of functions with multiple input values spread across processes.

%\subsubsubsection{PySpark Map-Reduce}
%\textcolor{blue}{
\paragraph{PySpark Map-Reduce}
%Map-Reduce is a programming model and processing technique designed to handle large-scale data processing tasks in a parallel and distributed manner.
Auto-labeling is a pixel-based process that is highly parallelizable.
Therefore, we also harnessed a Map-Reduce-based PySpark distributed platform - the Python API for Apache Spark - for further acceleration over a cluster.
The distributed auto-labeling procedure is implemented using the GCD cluster and PySpark. A cluster configuration consisting of four nodes was utilized, with one node designated as the master/executor and the remaining three nodes serving as worker/executor nodes. Each of the Intel N2 Cascade Lake computers is equipped with a total of four cores. 
%A {\it Map-Reduce}-based framework was created on the cloud, utilizing PySpark for the purpose of auto-labeling. This approach was adopted to achieve enhanced scalability and improved processing performance across multiple machines. 
First we read and load all the S2 image data into PySpark dataframe, which is a distributed collection of data capable of running on multiple machines.
We create a spark User-defined function (UDF) for our auto-labeling method.
Then we apply the {\it Map} transformation function using the auto-labeling UDF over S2 image Pyspark dataframe.
%These dataframes can pull from external databases, structured data files or existing resilient distributed datasets (RDDs).
Upon execution of the {\it Map} function, the entirety of the S2 data is transformed.
%, partitioned, and spread across a distributed file system. 
The transformed and distributed S2 data output from the map function is the input to the {\it Reduce} function. Finally, the {\it Reduce} function then collects all the auto-labeled S2 data from multiple machines as a result.
This is a linearly scalable approach for autolabeling.
%\textcolor{red}{check write about scaling linearly, pixel level parallelization pseudo code and images}
%}
\subsection{U-Net Model for Sea-Ice Classification}
U-Net is a fast training technique that utilizes data augmentation to better use the available annotated labeled data \cite{ronneberger2015u}. A good property of this convolutional network is that it can be trained end-to-end with a small training dataset. U-net is an efficient semantic segmentation model, and applying a multi-class U-Net in our sea ice dataset resulted in semantically segmented and classified sea ice. 

Our U-Net architecture is shown in Figure \ref{fig:unet_archi}. It consists of three parts: the first one is the contracting path (down-sampling focuses on what the feature is), the second one is the bottleneck, and the third one is the expansion path (up-sampling focuses on where the feature is). Each step of the contraction path consists of two consecutive 3x3 convolutional layers, each with a rectified linear unit (ReLU) \cite{hara2015analysis} followed by a 2x2 max-pooling \cite{nagi2011max} layer with stride 2. The bottleneck step is a single step similar to the down-sampling, except that there is no following max-pooling unit. After that comes the expansion path, where each step consists of (i) first an up-sampling of the feature map,  then (ii) a 2x2 convolution (up-convolution) that halves the number of feature channels, a concatenation with the proportionally cropped feature map from the contracting path, and after that (iii)  two 3x3 convolutions, each followed by a ReLU in the expansive path. Again, due to some loss of boundary pixels in every convolution layer, cropping is necessary. Finally, a 1x1 convolution layer transfers the input features to the desired activation function and the number of classes, which is 3 in our case for thick ice, thin ice, and water.

\begin{figure}[ht]
    \begin{framed}
        \centering
        \includegraphics[width=\textwidth]{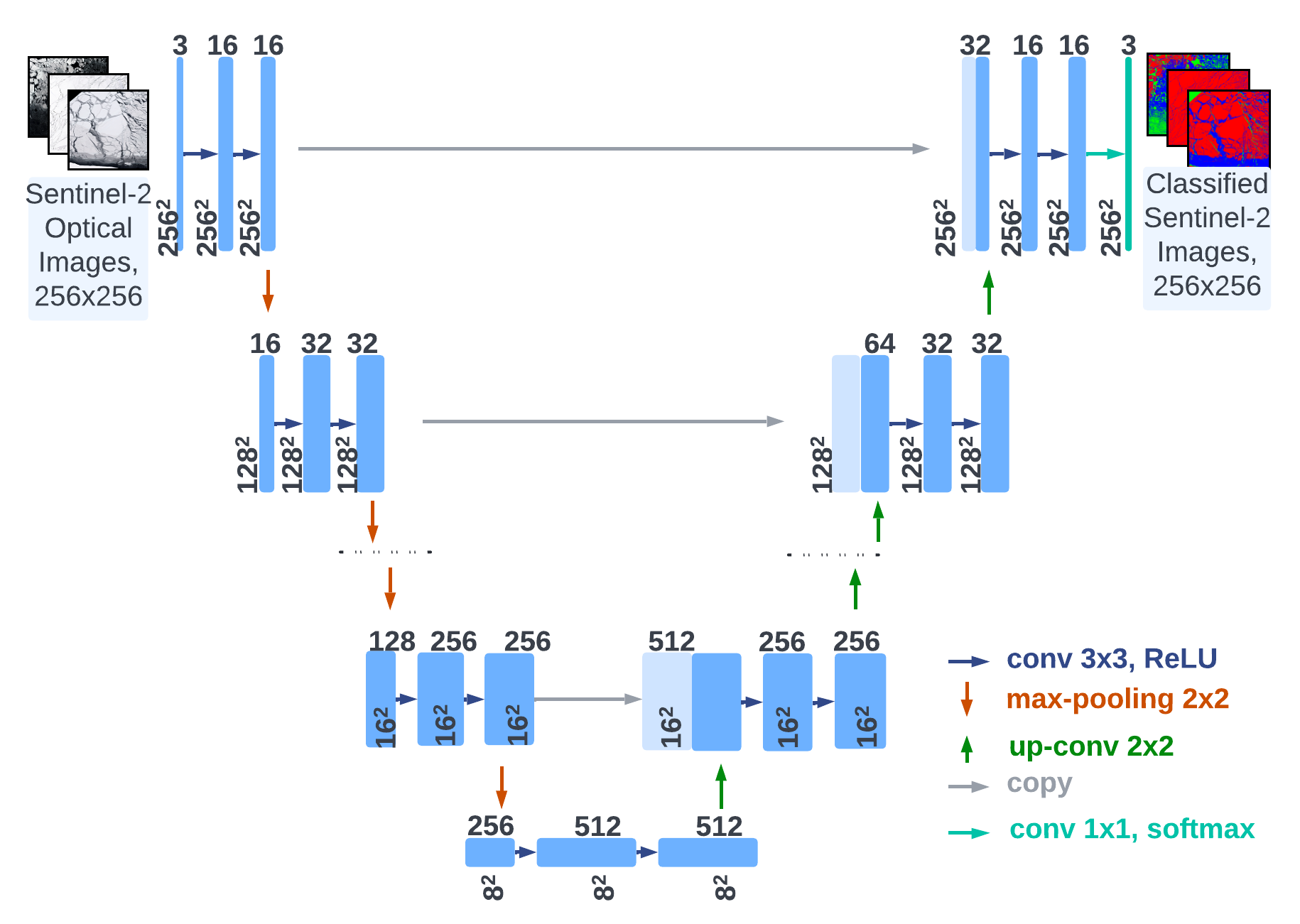}
    \end{framed}
    \caption{U-net model architecture. %\textcolor{red}{\tt [need larger fonts]}
    }
    \label{fig:unet_archi}
\end{figure}

\subsubsection{U-Net Training}

For our U-Net model, the input image size is 256x256 pixels. Our model has a total of 28 convolutional layers, including five downsampling steps, one bottleneck step, and five upsampling steps. We use the Adam optimizer for training and the categorical cross-entropy loss as a loss function for our multi-class model \cite{kingma2014adam}. Our model has some dropout layers in between our convolutional layers. The effect of dropout layers is to regularize the training to avoid overfitting \cite{srivastava2014dropout}. The Adam optimizer continuously improves the loss function on batches of the sample training image set. The optimization is done in epochs, with one epoch being reached when the entire dataset is provided to the neural network for optimization. 

%\textcolor{blue}{
\subsubsection*{Distributed U-Net Model Training using Horovod Framework}
Since training our U-Net model is computationally heavy, we want to focus on distributed training to make it more scalable without losing the accuracy of our model. We applied \textit{synchronized data parallelism} to scale the U-Net model training on multiple GPUs. 
%\textcolor{red}{talk about forward, backward, what it looks like, how it works, briefly about method for training and how it scales over multi node}
Instead of deploying one or multiple parameter servers, which are part of the built-in distribution strategy in TensorFlow, we opted to utilize Horovod \cite{sergeev2018horovod} for aggregating and averaging gradients across multiple GPUs. 
Horovod is an open-source distributed deep-learning training framework developed by Uber. 
%It is an open-source distributed training library for TensorFlow, Keras, PyTorch, and Apache MXNet. 
It enabled us to distribute the U-net model training over multiple GPUs. For efficient inter-GPU communication, it utilizes a ring-based all-reduce algorithm, which has been demonstrated to be bandwidth optimal and avoids system bottlenecks \cite{patarasuk2009bandwidth}.
MPI is used for coordinating between the processes in Horovod. The Open-MPI-based wrapper is utilized to execute the Horovod scripts.
To integrate our single-GPU implementation with the Horovod-based multiple-GPU distributed training,
we first initialize Horovod using \textit{hvd.init()} and assign a GPU to each of the TensorFlow processes. Then we wrap the TensorFlow optimizer with the Horovod optimizer using \textit{opt=hvd.DistributedOptimizer(opt)}. This Horovod optimizer handles gradient averaging using a ring-based all-reduce mechanism. Finally, we broadcast initial variable states from rank 0 to all other processes using \textit{hvd.callbacks.BroadcastGlobalVariablesCallback(0)}. 

% we followed the subsequent steps:
% \begin{enumerate}
%     \item Initialize Horovod using \textit{hvd.init()}.
%     \item Assign a GPU to each of the TensorFlow processes.
%     \item Wrap the TensorFlow optimizer with the Horovod optimizer using \textit{opt=hvd.DistributedOptimizer(opt)}. This Horovod optimizer handles gradient averaging using a ring-based all-reduce mechanism.
%     \item Broadcast initial variable states from rank 0 to all other processes using\\ \textit{hvd.callbacks.BroadcastGlobalVariablesCallback(0)}.
% \end{enumerate}

Figure \ref{fig:s2_unet_code} refers to training pseudo-code without and with Horovod integration.
Using Horovod, our model training became faster and more scalable without affecting the model's accuracy. 

\begin{figure}[htb]
    %\begin{framed}
        \centering
        \includegraphics[width=\linewidth]{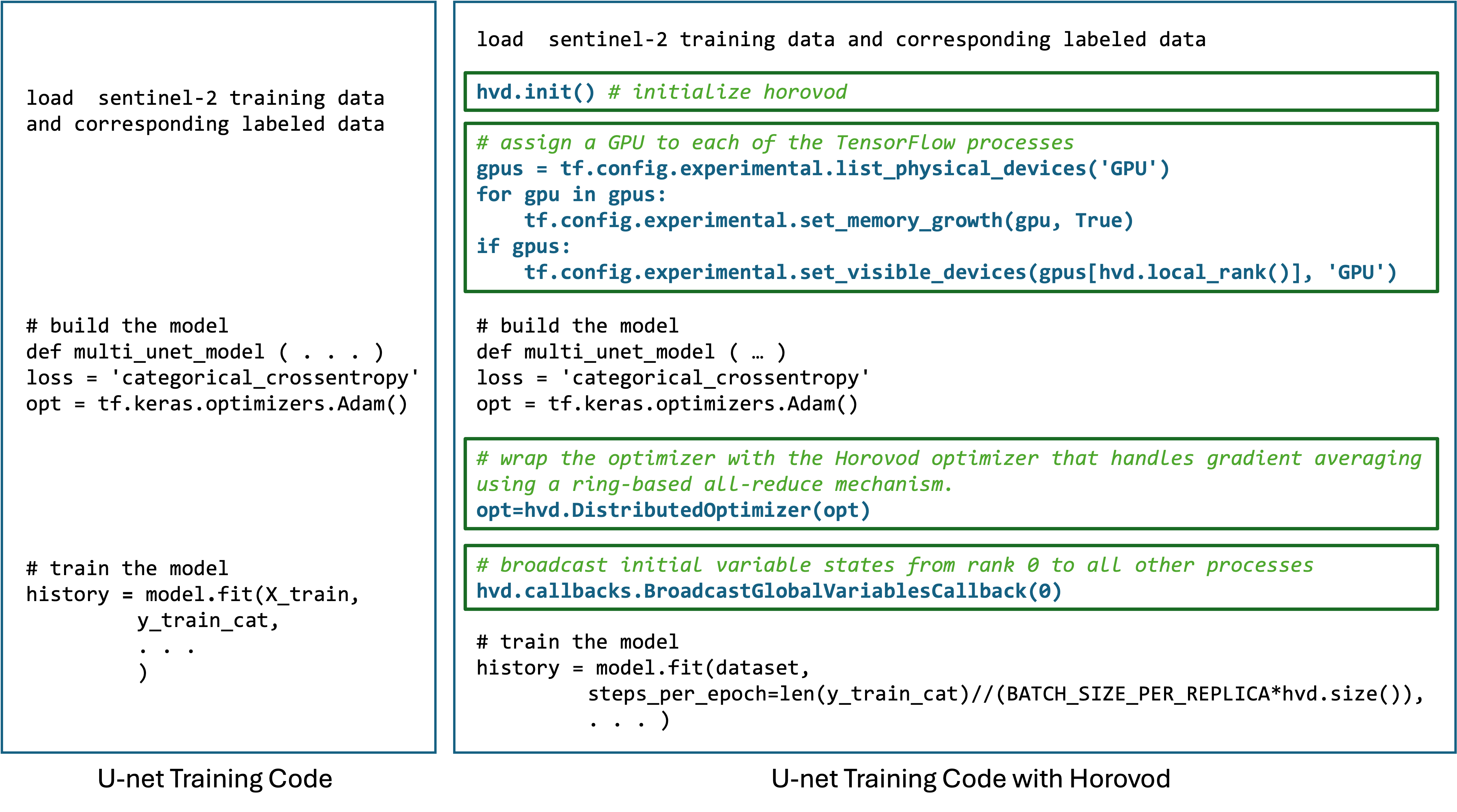}
    %\end{framed}
    \caption{U-Net training pseudo-code without and with Horovod
    }
    \label{fig:s2_unet_code}
\end{figure}

%}

\subsubsection{U-Net Model Inferencing}
A high-precision pre-existing U-Net model has been trained for the purpose of classifying sea ice using the S2 image dataset.
%Our proposed approach involves the implementation of distributed inferencing for our model over a range of heterogeneous machines, with the aim of improving scalability.
\begin{figure}[htb]
    \begin{framed}
        \centering
        \includegraphics[width=\linewidth]{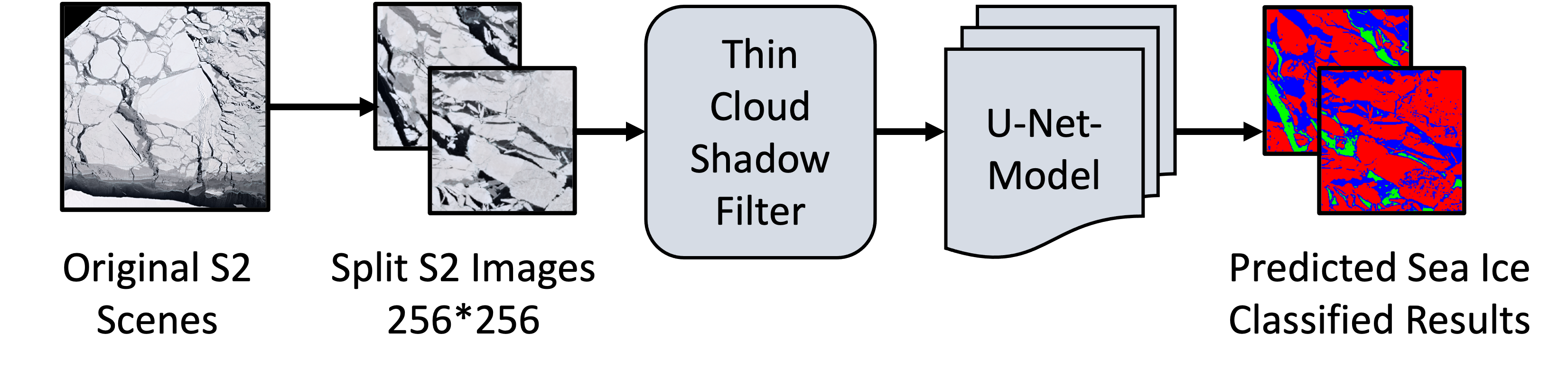}
    \end{framed}
    \caption{Workflow for Sea Ice Classification employing cloud-shadow filter and U-net model Inferencing.
    }
    \label{fig:s2_infer}
\end{figure}
In the inferencing workflow of the U-net model, as depicted in Figure \ref{fig:s2_infer}, the initial step involves the acquisition of the original S2 big scenes. The resolution of the R, G, and B bands in these images is 10 meters. The process involves dividing the huge images into smaller 256*256 images, which will then be utilized as input for the U-net model during the inferencing phase. Prior to providing the U-net model, our thin cloud and shadow filter technique is employed to effectively exclude thin clouds and cloud shadows from the image, hence enhancing the accuracy of the inference results. Subsequently, the filtered image is forwarded to the model in order to obtain the predicted image that classifies sea ice.

%The Python multiprocessing library and thread pool were employed to enhance the scalability of the inferencing process. Speedup was observed on the multicore machines as well. However, additional research is required to enhance this scaling strategy. The objective of our study is to employ spark-based distributed inferencing as a means to enhance the scalability and efficiency of inferencing over several workstations.

\section{Experiments and Results}\label{sec:expr_res}
%\section{Experiments and Results}\label{sec:expr_res}
\subsection{Experimental Setup and Evaluation Metrics}
We collected S2 sea ice optical RGB band data using the GEE. We collected 66 large scenes from the Ross Sea region in the Antarctic. First, we split the large S2 scenes into 4224 images, each with 256x256 pixels. Then, we derived the ground truth/manually labeled data. 
%After that, for cloud and shadow, we divided the cloudy-shadowy images from the cloud and shadow-free images. 
We separated the images into cloudy-shadowy images and cloud-shadow-filtered images.
Finally, we filtered thin clouds and shadows using OpenCV library image transformation techniques on the cloudy-shadowy data. For the color-based segmentation, as it is a color-limit-based approach, we used the OpenCV library to process this computation. 
%Then we applied efficient parallel implementation for this color-segmentation using the python multiprocessing library to use the pool of processes. 
%For this experiment, we have used a 2 GHz Quad-Core Intel Core i5 processor with eight cores (four physical and four virtual); details are shown in table \ref{tab:conf}.

We have applied two techniques for our auto-labeling scaling and parallelization: the Python multiprocessing approach for a single node and the PySpark map-reduce-based approach for multiple nodes. For Python multiprocessing experiments, we have used a 2 GHz Quad-Core Intel Core i5 processor with four physical cores with hyperthreading. We have utilized GCD service for PySpark-based experiments. There, we have used a cluster of four nodes with one master node and three worker nodes. Each of the Intel N2 Cascade Lake computers was equipped with a total of four cores.

As for the U-Net models, first, we divided the dataset into 80\% training dataset and 20\% test dataset. Then, we organized the data into batches for the U-Net models using dataloader. We have used the Adam optimizer, batch sizes of 16, 32, and 64, dropouts of 0.1, 0.2, and 0.3 in different convolutional layers, and epochs of 50, 70, and 100 to observe the changes. Our U-Net models have a batch size of 32, and the number of epochs is 50 for the results reported in the next section. 
%This U-net training is computationally heavy than the color segmentation. For this, we have used Intel Xeon(R) Silver 4210R 2.40GHz × 20 processors and NVIDIA Quadro RTX 5000 GPU as shown in the table \ref{tab:conf}.
Since this U-net training is computationally heavy, we have applied Horovod-based distributed training. For this, we have utilized an NVIDIA DGX A100 machine with dual CPUs, each with four A100 GPUs.

%\subsubsection{Evaluation Metrics}
To validate the results of our algorithm, the following evaluation metrics over the validation dataset are computed:
\begin{itemize}
    %\item Thin cloud and shadow removal:\\We have 
    \item Classification Accuracy, Precision, Recall, and F1 score:
    %The overall accuracy is the ratio of the correctly predicted samples over all the samples in the validation set. 
    %For measuring the overall accuracy of the color-seg model, we measured SSIM (structural similarity index) to find similarity accuracy between model output with the target image. 
    We compute the accuracy, precision, recall, and F1 score to obtain a comprehensive and balanced evaluation of the model's performance.
    For the two U-Net models(one trained on a manually labeled dataset and another on the auto-labeled dataset), we evaluate the models using a ground truth validation dataset to find the overall classification accuracy of these two models.  
    
    \item Confusion Matrix:
    For our segmentation model evaluation, we also construct a confusion matrix \cite{ting2017confusion}. The number of samples predicted in category A over the number of samples in category B is specified as an element of the matrix in row A and column B. A complete classification would result in a diagonal confusion matrix, with 100\% on the diagonal and 0\% in the rest of the matrix. Each column adds up to a total of 100\%. This helps understand the model accuracy for classifying each sea ice type individually.
    
\end{itemize}

\subsection{Auto-labeling}

\subsubsection{Auto-labeling Speedup}
Auto-labeling speedup results for the Python multiprocessing-based approach and PySpark-based approach are as follows,

\paragraph{Python Multiprocessing Performance}
Parallelization was employed throughout the entire process of automatically labeling S2 images, utilizing thin clouds and shadow-filtered color-based segmentation techniques. This facilitated an acceleration in the rate at which the procedure was executed.
In a parallel computing environment, we saw a significant speedup of up to 4.5 times compared to a sequential execution. This improvement allowed us to analyze a total of 4224 images within 3.89 seconds. The speedup and parallel processing time for this can be observed in Table \ref{tab:parallel} and Figure \ref{fig:para_chart}, respectively.

\begin{table}[htb]
\centering
\caption{Python Multiprocessing base Auto-labeling.}
    \label{tab:parallel}
\begin{tabular}{ |>{\centering\arraybackslash}p{0.22\linewidth}||>{\centering\arraybackslash}p{0.18\linewidth}||>{\centering\arraybackslash}p{0.17\linewidth}||>{\centering\arraybackslash}p{0.20\linewidth}|}
%\begin{tabular}{|c||c||c||c|}
\hline
\multirow{2}{*}{\begin{tabular}[c]{@{}c@{}}\textbf{No. of  Processes}\end{tabular}} & \textbf{Parallel time (s)} & \textbf{Sequential time (s)}       & \textbf{Speedup}     \\ \cline{2-4} 
                                                                             & \textbf{Tp}            & \textbf{Ts}                    & \textbf{S = Ts / Tp} \\ \hline \hline
1                                                                            & 17.40          & \multirow{5}{*}{\textbf{17.40}} & 1.0           \\ \cline{1-2} \cline{4-4} 
2                                                                            & 8.89          &                       & 2.0           \\ \cline{1-2} \cline{4-4} 
4                                                                            & 4.69          &                       & 3.7         \\ \cline{1-2} \cline{4-4} 
6                                                                            & 4.10           &                       & 4.2         \\ \cline{1-2} \cline{4-4} 
8                                                                            & \textbf{3.89}          &                       & \textbf{4.5}         \\ \hline
\end{tabular}
\end{table}

\begin{figure}[htb]
    %\begin{framed}
        \centering
        \frame{\includegraphics[width=1.0\linewidth]{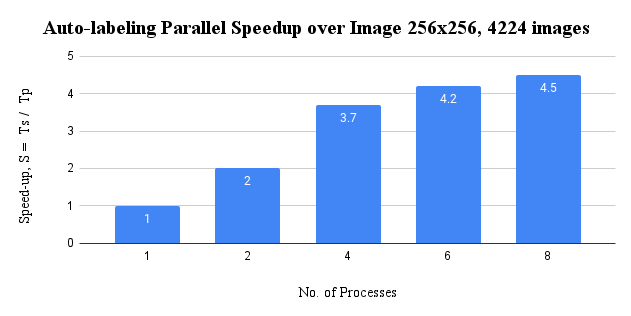}}
    %\end{framed}
    \caption{Parallel execution speedup for color-segmentation-based auto-labeling on a 4-core machine. }
    \label{fig:para_chart}
\end{figure}

\begin{table}[htb]
\centering
\caption{PySpark-based auto-labeling scalability over Google Cloud.}
\label{tab:parallelspark}
\begin{tabular}{ |>{\centering\arraybackslash}p{0.115\linewidth}||>{\centering\arraybackslash}p{0.06\linewidth}||>{\centering\arraybackslash}p{0.06\linewidth}||>{\centering\arraybackslash}p{0.06\linewidth}||>{\centering\arraybackslash}p{0.08\linewidth}||>{\centering\arraybackslash}p{0.11\linewidth}||>{\centering\arraybackslash}p{0.11\linewidth}|}
%\begin{tabular}{|c|c|c|c|c|c|c|}
\hline
\textbf{Executors} & \textbf{Cores} & \textbf{Load Time} & \textbf{Map Time} & \textbf{Reduce Time} & \textbf{Speed-up Load} & \textbf{Speed-up Reduce} \\ \hline
1                  & 1              & 108                & 0.4               & 390                  & 1                      & 1                        \\ \hline
1                  & 2              & 58                 & 0.4               & 174                  & 1.86                   & 2.24                     \\ \hline
1                  & 4              & 33                 & 0.3               & 72                   & 3.27                   & 5.42                     \\ \hline
2                  & 1              & 56                 & 0.3               & 156                  & 1.93                   & 2.5                      \\ \hline
2                  & 2              & 31                 & 0.3               & 84                   & 3.48                   & 4.64                     \\ \hline
2                  & 4              & 19                 & 0.3               & 41                   & 5.68                   & 9.51                     \\ \hline
4                  & 1              & 31                 & 0.2               & 78                   & 3.48                   & 5                        \\ \hline
4                  & 2              & 17                 & 0.2               & 39                   & 6.35                   & 10                       \\ \hline
4                  & 4              & 12                 & 0.3               & 24                   & \textbf{9}                      & \textbf{16.25}                    \\ \hline
\end{tabular}
\end{table}

\paragraph{PySpark Performance}
The GCD service was utilized alongside the PySpark framework for thin cloud and shadow-filtered autolabeling. This technique was applied for the annotation of S2 data that was subsequently used for training deep learning models. A speedup of up to 16.25 times has been attained in the execution of this workflow, as illustrated in Table \ref{tab:parallelspark}. Additionally, there was a significant improvement in image loading speed when utilizing numerous machines, with a maximum speedup of up to 9 times. 

%\paragraph*{Discussion}
%\textcolor{red}{why pyspark better comparison}
The auto-labeling of S2 images is highly scalable due to independent pixel processing, albeit fine-grained. This can easily be parallelized using Python multiprocessing on a single machine with a speedup of 4.5x. However, PySpark (using the map-reduce framework) can be utilized to parallelize and scale the auto-labeling of the S2 imagery on different architectures. Along with a single multi-core machine, it is scaled over multiple heterogeneous machines in a GCD cluster with 9.0x data loading and 16.25x map-reduce processing speedup. Since the PySpark-based approach supports larger clusters for distributing the auto-labeling procedure, it points to a potential for scaling over much larger data in future.

\subsubsection{Auto-labeling Accuracy}
%Auto-labeling based on color segmentation is very light in terms of processing time and resource demand. 
For the original S2 data and the S2 data with clouds and shadows filtered out, we respectively achieved 89\% and 99.64\% Structural Similarity Index (SSIM) \cite{ma2017multi} precision over the manually labeled data. Some sample results of the color segmentation approach are shown 
in Figure \ref{fig:colsegsample}.
\begin{figure}[ht]
    \begin{framed}
        \centering
        \begin{subfigure}[b]{0.45\textwidth}
            \centering
            \frame{\includegraphics[width=\textwidth]{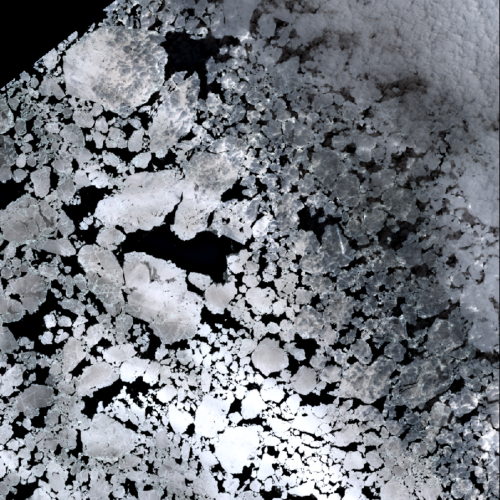}}
            \caption{}
            \label{fig:s2ori}
        \end{subfigure}
        \hfill
        \begin{subfigure}[b]{0.45\textwidth}
            \centering
            \frame{\includegraphics[width=\textwidth]{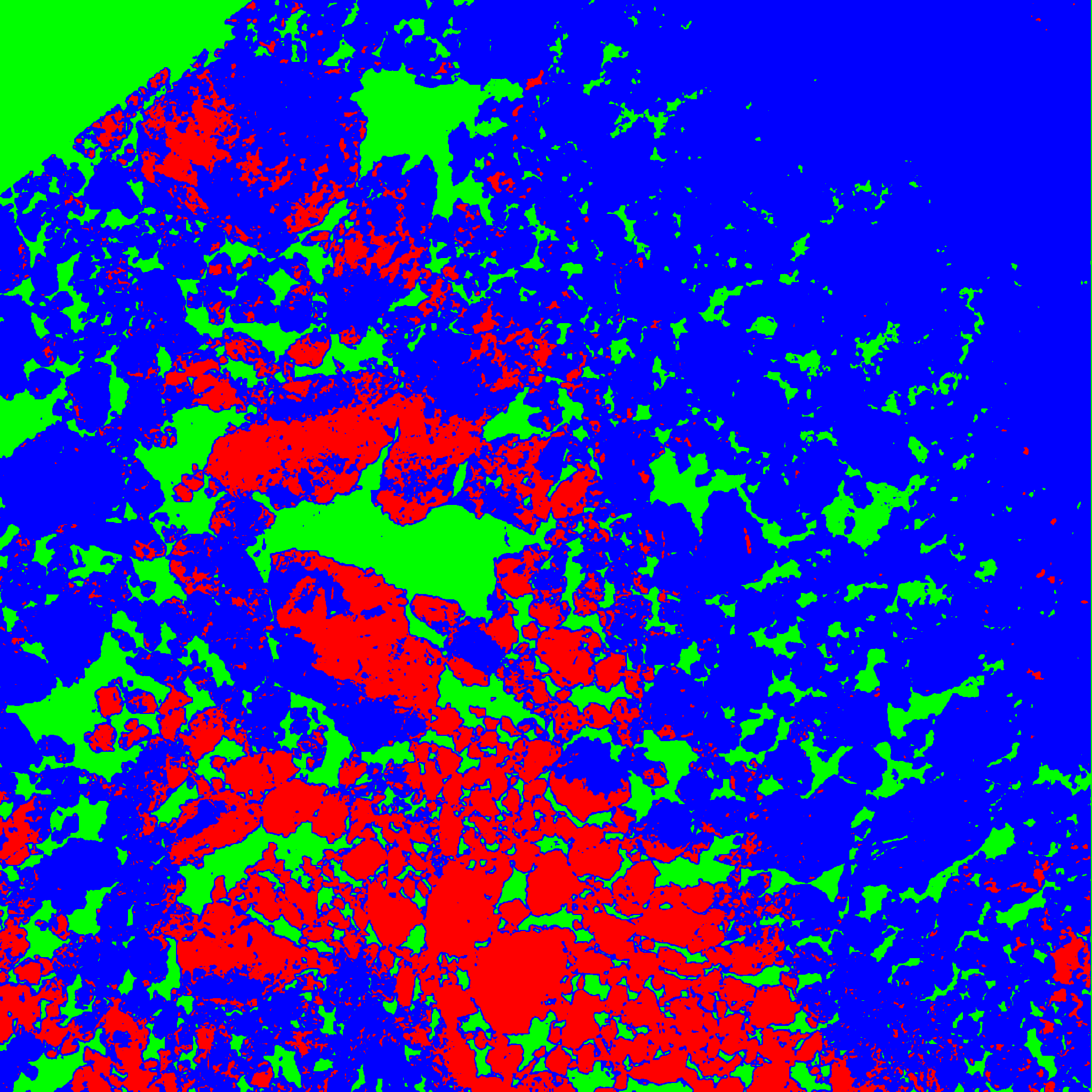}}
            \caption{}
            \label{fig:s2colori}
        \end{subfigure}
        %\par\medskip
        \begin{subfigure}[b]{0.45\textwidth}
            \centering
            \frame{\includegraphics[width=\textwidth]{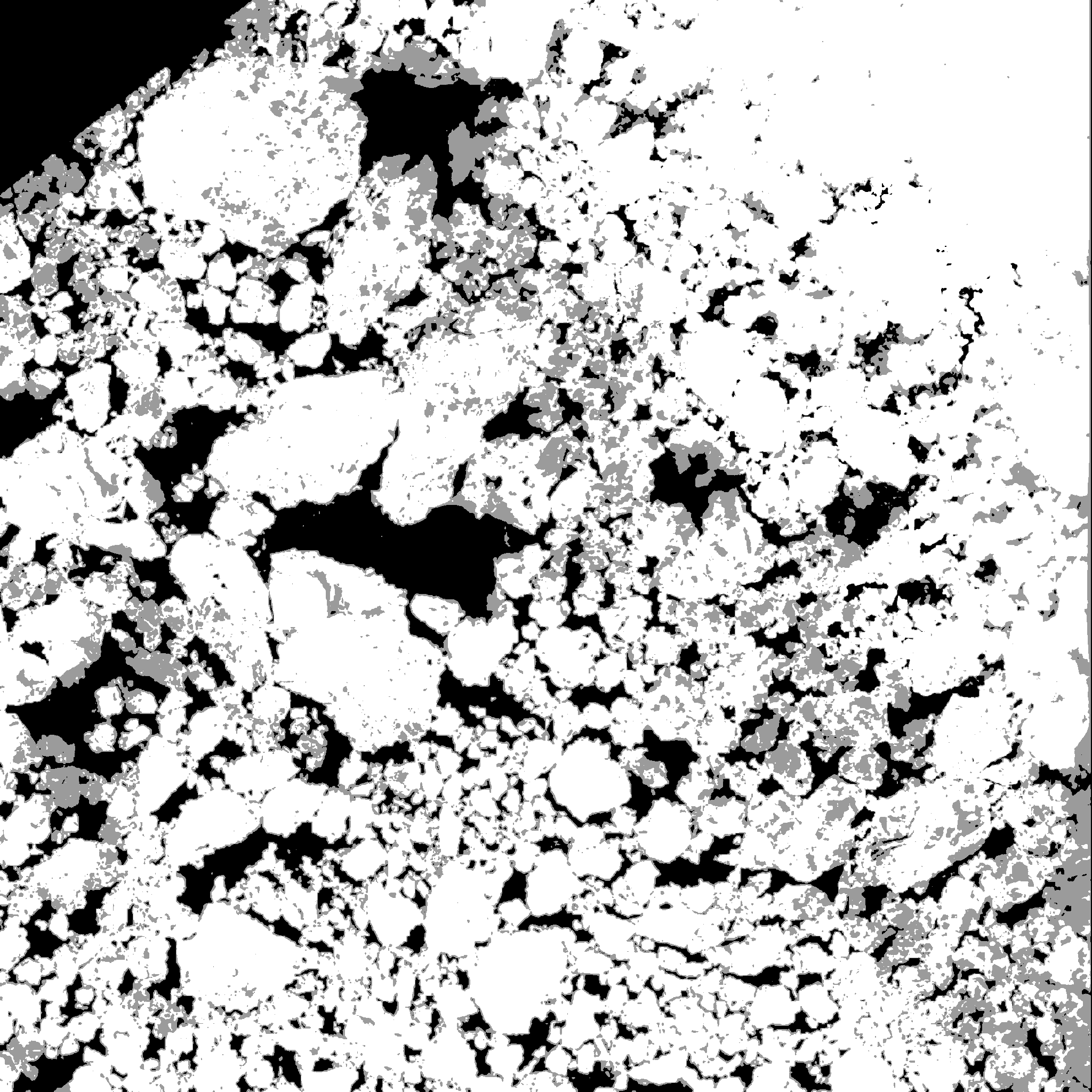}}
            \caption{}
            \label{fig:s2ref}
        \end{subfigure}
        \hfill
        \begin{subfigure}[b]{0.45\textwidth}
            \centering
            \frame{\includegraphics[width=\textwidth]{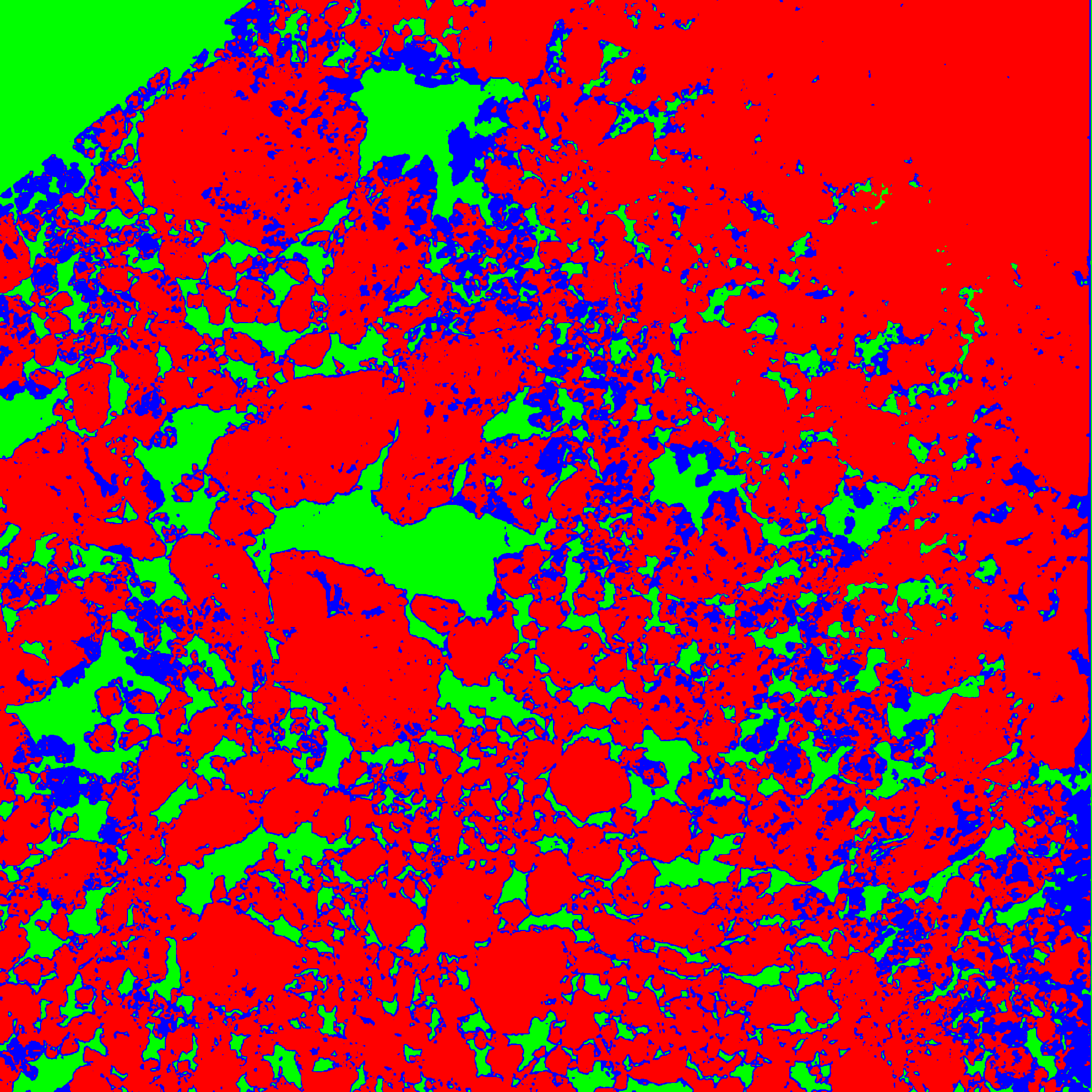}}
            \caption{}
            \label{fig:s2colref}
        \end{subfigure}
        %\par\medskip
        \begin{subfigure}[b]{0.9\textwidth}
            \centering
            \includegraphics[width=\textwidth]{figures/legends.png}
        \end{subfigure}       
    \end{framed}
    \caption{Color-segmentation based auto-labeling, (a) Sentinel-2 thin cloudy and shadowy image and (b) color-segmented version of (a) with erroneous segmentation in the cloudy/shadowy areas; (c) represents (a)'s corresponding thin shadow and cloud filtered image and (d) represents the color-segmented version of (c).}
    \label{fig:colsegsample}
\end{figure}

We note that the color limits for color-segmentation are not independent of different regions and seasons. For the partial night season of the Antarctic, we had to change the color threshold brightness values manually to label those data to regain accuracy (not reported here). Likewise, the same color limits may not work for different regions of sea ice labeling, and a manual color limit setup may be needed in those cases. On the other hand, we expect our machine model to be robustly trained on the auto-labeled data over various seasons and regions of the poles.

\begin{table}[htb]
\caption{Distributed U-Net model training using Horovod framework on DGX A100 cluster.}
\label{tab:u_net_training_horovod}
\begin{tabular}{ |>{\centering\arraybackslash}p{0.10\linewidth}||>{\centering\arraybackslash}p{0.12\linewidth}||>{\centering\arraybackslash}p{0.25\linewidth}||>{\centering\arraybackslash}p{0.12\linewidth}||>{\centering\arraybackslash}p{0.12\linewidth}|}
%\begin{tabular}{|c|c|c|c|}
\hline
\textbf{No. of GPUs} & \textbf{Time (s)} & \textbf{Time (s)/Epoch} & \textbf{Data/s} & \textbf{Speedup}     \\ \hline
% 1           & 1657.65     & 34.0010                                 & 95.92      & 1.00    \\ \hline
% 2           & 923.64  & 18.0011                                & 178.32     & 1.79    \\ \hline
% 4           & 581.64 & 11.0014                                & 287.23     & 2.85    \\ \hline
% 6           & 516.09 & 10.0018                                & 329.49     & 3.21    \\ \hline
% 8           & 479.91 & 9.0022                                & 360.48     & 3.45    \\ \hline
1          & 280.72                        & 5.5                                 & 585.88                       & 1.00    \\ \hline
2          & 142.98                        & 2.778                               & 1160.81                      & 1.96    \\ \hline
4          & 74.09                         & 1.45                                & 2229.56                      & 3.79    \\ \hline
6          & 51.56                         & 0.97                                & 3330.03                      & 5.44    \\ \hline
8          & 38.91                         & 0.79                                & 4248.56                      & 7.21   \\ \hline
\end{tabular}
\end{table}

\begin{figure}[htb]
    %\begin{framed}
        \centering
        \begin{subfigure}[b]{0.24\textwidth}
            \centering
            \frame{\includegraphics[width=\textwidth]{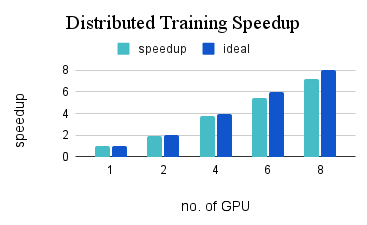}}
            \caption{}%{Distributed Training Speedup}
            \label{fig:distributed_training_speedup}
        \end{subfigure}
        %\hfill
        \begin{subfigure}[b]{0.24\textwidth}
            \centering
            \frame{\includegraphics[width=\textwidth]{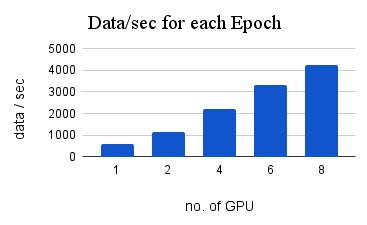}}
            \caption{}%{Data/sec for each Epoch}
            \label{fig:data_sec_per_epoch}
        \end{subfigure}
        \par\medskip
        %\hfill
        \begin{subfigure}[b]{0.24\textwidth}
            \centering
            \frame{\includegraphics[width=\textwidth]{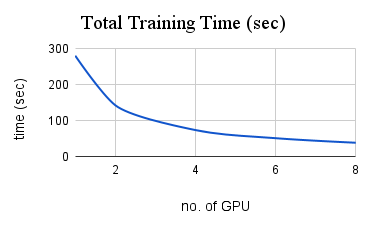}}
            \caption{}%{Total_Training Time(sec)}
            \label{fig:total_train_time}
        \end{subfigure}
        %\hfill
        \begin{subfigure}[b]{0.24\textwidth}
            \centering
            \frame{\includegraphics[width=\textwidth]{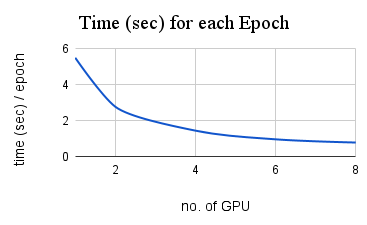}}
            \caption{}%{Time(sec) for each Epoch}
            \label{fig:time_per_epoch}
        \end{subfigure}
        %\hfill
        %\medskip
    %\end{framed}
    \caption{Distributed U-Net model training via Horovod framework, 
    (a) distributed training speedup, 
    (b) data processed per second for each epoch, 
    (c) total training time over multiple GPUs, and 
    (d) time for each epoch.}
    \label{fig:horovod_training_chart}
\end{figure}

\subsection{U-Net Model Training}

\subsubsection{U-net Model Distributed Training Speedup}
Table \ref{tab:u_net_training_horovod} shows the scaled and distributed U-net model training results. We trained our model in the DGX A100 cluster using the Horovod framework. We calculated the time for our Horovod-based U-Net model training in 1, 2, 4, 6, and 8 GPU setups and a batch size of 32 per node. The training time was reduced from 280.72s for 1 GPU to 38.91s for 8 GPU, gaining up to 7.21x speedup. We have trained our model for 50 epochs, and for each epoch, we achieved up to 4248.56 image data/s throughput within 0.79s on 8 GPUs compared to 585.88 image data/s throughput with 5.5s on a single GPU. Figure \ref{fig:horovod_training_chart} shows the performance results of distributed model training via Horovod over multiple GPUs. 

Here, we observe that the distributed training speedup and the throughput growth rate are almost close to linear, which is ideal with the increased number of GPUs.
On the other hand, the total training time and the time per epoch decreasing rate are high initially; however, eventually, they slow down with the increased number of GPUs. 
%This happens due to data starvation on the GPU side. 
%We notice that we are not getting ideal linear speedup and throughput or reduced total training time and time per epoch.
During training, the bottleneck arises from data preprocessing and subsequent batch preparation, resulting in GPU starvation. As a result, we are not achieving optimal speedup or throughput performance.

% The distributed training speed-up slows down after 4 GPUs due to the communication overhead across two groups of GPUs, as the DGX cluster is a dual CPU with 4 GPUs each.
% - based on paper,
% single node GPU vs multinode node
% communication
% 4k 8k 12k data GPU utilization 
%we know why its happening
%not scaling linearly at all since it has to copy the weights to all cards after each optimizer step which takes a lot of memory read/write operations.
%

\subsubsection{U-Net Model Accuracy}

The accuracy comparison of the U-Net-Man and U-Net-Auto is presented in Table \ref{tab:val_acc_comp} for S2 images without thin cloud and shadow filter and S2 images with thin cloud and shadow filter. The U-Net-Man has an accuracy of 91.39\% for original S2 images, and the U-Net-Auto has an accuracy of 90.18\%. 
The U-Net-Man achieved 91.11\% precision, 91.12\% recall, and 91.10\% F1 score. On the other hand, U-Net-auto achieved 91.14\% precision, 91.05\% recall, and 91.10\% F1 score.
After applying the thin cloud and shadow filter to the S2 images, the accuracy of the U-Net-Man and U-Net-Auto increases to 98.40\% and 98.97\%, respectively. For the thin cloud and shadow-filtered images, U-Net-Man achieved 98.35\% precision, 98.35\% recall, and 98.38\% F1 score; U-Net-auto achieved 98.88\% precision, 91.87\% recall, and 91.89\% F1 score.

\begin{table}[htb]
\caption{U-Net models sea ice classification accuracy over Sentinel-2 Antarctic summer datasets. }
\label{tab:val_acc_comp}
\begin{tabular}{ |>{\centering\arraybackslash}p{0.23\textwidth}||>{\centering\arraybackslash}p{0.08\textwidth}||>{\centering\arraybackslash}p{0.08\textwidth}|>{\centering\arraybackslash}p{0.00\textwidth}}
\hline
\textbf{Dataset}                                              & \textbf{U-Net Man} & \textbf{U-Net Auto} \\ \hline \hline
Original S2 images            & 91.39\%                  & 90.18\%                  \\ \hline
S2 images with thin cloud and shadow filtered              & 98.40\%                  & 98.97\%                   \\ \hline
\end{tabular}
\end{table}

Thus, the U-Net-Man and U-Net-Auto have very similar accuracy for classifying sea ice cover. However, after thin cloud and shadow filter on S2 images, the classification accuracy increases for both U-Net-Man and U-Net-Auto by about 7\% and 8\%, respectively.

For elaborate comparison, we further divided the S2 validation dataset into (i) more than about 10\% cloud and shadow cover and (ii) less than about 10\% cloud and shadow cover datasets. S2 sea ice classification validation accuracy comparison of U-Net-Man and U-Net-Auto over these two different datasets are represented in Table \ref{tab:val_acc_comp2}. We also included the accuracy of thin cloud-shadow-filtered original images along with the original images.

\begin{table}[htb]
\caption{Sentinel-2 sea ice classification validation accuracy comparison over increasing cloud/shadow coverage.}
%. Based on cloud and shadow cover we divided the datasets into more than about 10\% cloud and shadow cover and less than about 10\% cloud and shadow cover. Besides the accuracy over original S2 images we also added the accuracy over thin cloud and shadow filtered images for both of these datasets.}
\label{tab:val_acc_comp2}
%\begin{tabular}{|ll|l|l|}
\begin{tabular}{ |>{\centering\arraybackslash}p{0.3\textwidth}||>{\centering\arraybackslash}p{0.06\textwidth}||>{\centering\arraybackslash}p{0.06\textwidth}|>
{\centering\arraybackslash}p{0.06\textwidth}|>
{\centering\arraybackslash}p{0.00\textwidth}}
\hline
%\multicolumn{1}{|p{3cm}||}{\centering \textbf{\\Model}}&
\multicolumn{2}{|p{4cm}|}{\centering \textbf{Dataset}}                                                                        & \textbf{U-Net-Man} & \textbf{U-Net-Auto} \\ \hline
\hline
\multicolumn{1}{|p{3.25cm}|}{\multirow{2}{0.19\textwidth}{More than about 10\% cloud and shadow cover}} & original images & 88.74\%         & 79.91\%           \\ \cline{2-4} 
\multicolumn{1}{|p{1cm}|}{}                                                             & filtered images & 98.91\%         & 99.28\%          \\ \hline
\hline
\multicolumn{1}{|p{3.25cm}|}{\multirow{2}{0.19\textwidth}{Less than about 10\% cloud and shadow cover}} & original images & 92.27\%         & 93.60\%          \\ \cline{2-4} 
\multicolumn{1}{|p{1cm}|}{}                                                             & filtered images & 98.23\%         & 98.87\%          \\ \hline
\end{tabular}
\end{table}

%unet confusion matrix figure
\begin{figure*}[!ht]%[ht]
    %\begin{framed}
        \centering
        \frame{\includegraphics[width=0.9\textwidth]{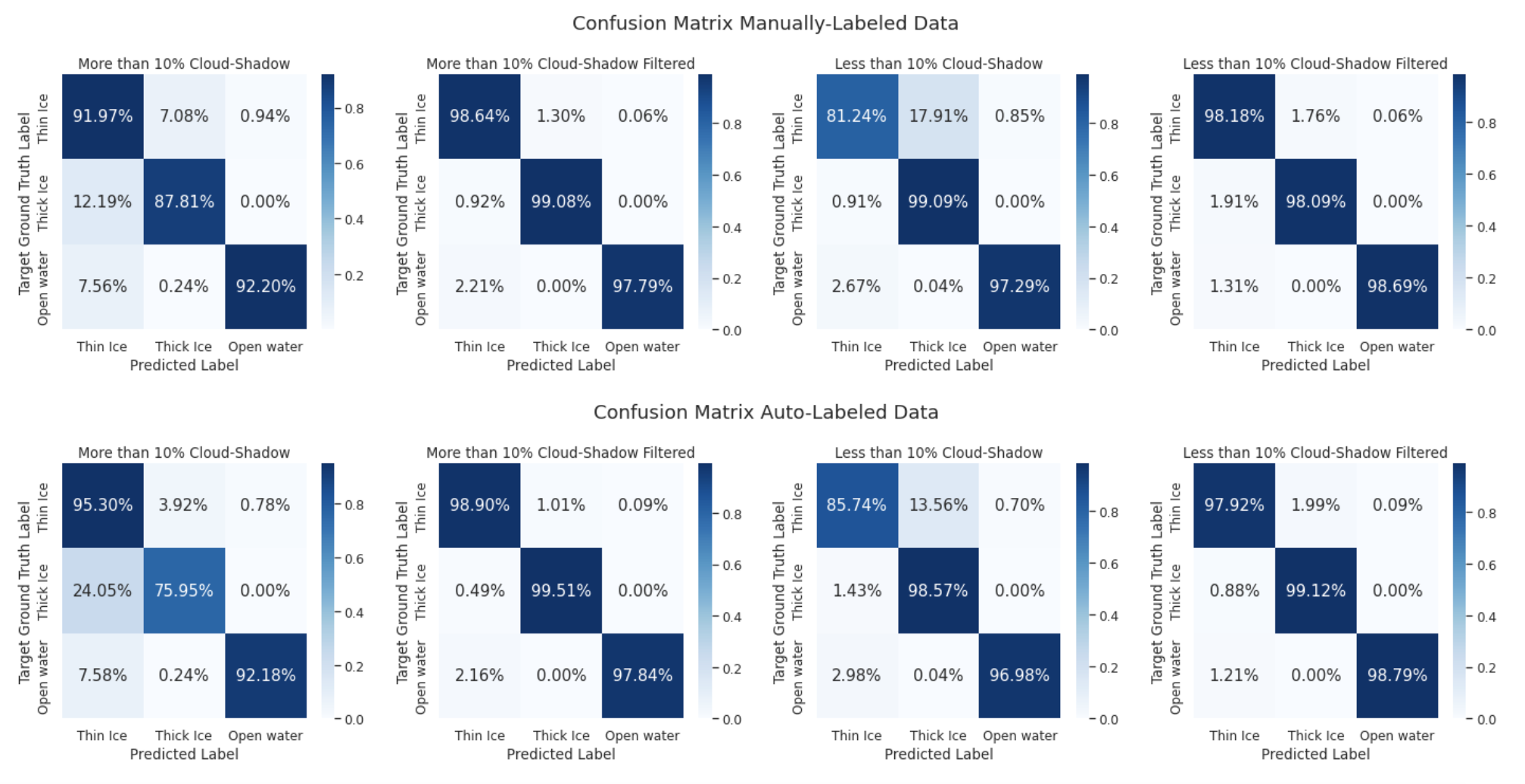}}     
    %\end{framed}
    \caption{Confusion Matrix of manually labeled and auto-labeled U-net model for thin ice, ice, and open water accuracy over cloudy-shadowy, cloud-shadow-removed, and cloud-shadow-free.}
    \label{fig:confusion_matrix}
\end{figure*}
%%%%

In terms of this cloudy and shadowy dataset, U-Net-Man has a better sea ice classification accuracy of 88.74\% compared to U-Net-Auto with 79.91\%. However, after applying our thin cloud and shadow filter, the accuracy of our U-Net-Auto increased to 99.28\%, which is nearly a 20\% increment. On the other hand, U-Net-Man increased to 98.91\% with a 10\% increment.
Again, on less than about 10\% cloud and shadow cover original and thin cloud and shadow filtered S2 images, U-Net-Man has 92.27\% and 98.23\% accuracy, respectively, whereas U-Net-Auto has 93.60\% and 98.87\%. For this less cloudy and shadowy dataset, U-Net-Man and U-Net-Auto have similar sea ice classification accuracy, increasing by over 5\% for both models on thin cloud-shadow-filtered images. We do not handle thick clouds or shadows here and plan to do that in the future.

%unet pred figure
\begin{figure}[ht]%[!hb]
    %\begin{framed}
        \centering        
        \begin{subfigure}[b]{0.18\textwidth}
            \centering
            \frame{\includegraphics[width=\textwidth]{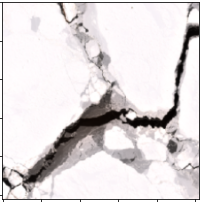}}
            \caption{Original image}
            \label{fig:pred-ori}
        \end{subfigure}
        %\hfill
        \begin{subfigure}[b]{0.18\textwidth}
            \centering
            \frame{\includegraphics[width=\textwidth]{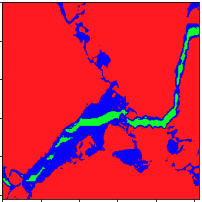}}
            \caption{Ground Truth }
            \label{fig:pred-lbl}
        \end{subfigure}
        
        \par\medskip
        
        \begin{subfigure}[b]{0.18\textwidth}
            \centering
            \frame{\includegraphics[width=\textwidth]{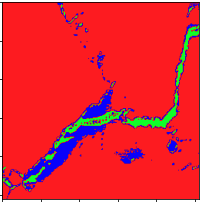}}
            \caption{U-Net-Man Pred}
            \label{fig:pred-man}
        \end{subfigure}
        %\hfill
        \begin{subfigure}[b]{0.18\textwidth}
            \centering
            \frame{\includegraphics[width=\textwidth]{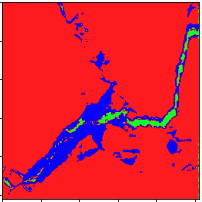}}
            \caption{U-Net-Auto Pred}
            \label{fig:pred-auto}
        \end{subfigure}
        \medskip
        
        \begin{subfigure}[b]{0.4\textwidth}
            \centering
            \includegraphics[width=\textwidth]{figures/legends.png}
        \end{subfigure}     
    %\end{framed}
    \caption{S2 original image and corresponding manually labeled ground truth image compared to U-Net-Man and U-Net-auto model predictions.}
    \label{fig:validation-auto-man}
\end{figure}

We examine the confusion matrix of thin ice, ice, and open water for the U-Net models with epoch 50 in Figure \ref{fig:confusion_matrix}. It includes the individual accuracy of each class, %with thin ice, ice, and open water 
along the diagonal of the matrix. It also indicates that the thin cloud and shadow filtering improves the individual accuracy of each class along with the overall accuracy shown in Tables \ref{tab:val_acc_comp} and \ref{tab:val_acc_comp2}.
%%%%%
For both U-net models, the accuracy of thick ice, thin ice, and open water is similar (about 98\%) in confusion matrices based on thin cloud and shadow-filtered data. 
However, for original data without the thin cloud and shadow filter, when there is more than 10\% cloud and shadow in the image, 12.19\% (U-Net-Man) and 24.05\% (U-Net-Auto) of thick ice are classified as thin ice due to the shadows.
Due to the clouds, the models classify 7.08\% (U-Net-Man) and 3.92\% (U-Net-Auto) thin ice as thick ice; 7.56\% (U-Net-Man) and 7.58\% (U-Net-Auto) open water as thin ice.
In less than 10\% cloudy and shadowy data, thick ice has 99.09\% (U-Net-Man) and 98.57\% (U-Net-Auto) accuracy.  
Due to some presence of thick and thin clouds in the data, thin ice is classified as thick ice by 17.91\% (U-Net-Man) and 13.56\% (U-Net-Auto), and 2.67\% (U-Net-Man) and 2.98\% (U-Net-Auto) open water as thin ice.

%computational complexity/wall clock numbers of the entire training pipeline in the future iteration of the paper.
Our color-segmented auto-labeled S2 data preparation takes 349.26 seconds for 66 large S2 scenes of 2048x2048 pixels. This time includes our thin cloud and shadow filter method followed by color segmentation to label the S2 images. 
Then, for training our U-Net models, we split the original 66 scenes and their corresponding color-segmented thin cloud and shadow-filtered auto-labeled images into 4224 256x256 pixel images.
%The wall clock numbers of the entire training pipeline for the U-Net-Man model and U-Net-Auto are 1105.88 and 1178.04 seconds, respectively. 
%The whole experiment was executed in a Ubuntu 20.4 system with Intel Xeon(R) Silver 4210R 2.40GHz × 20 CPU, 64GB RAM, and NVIDIA Quadro RTX 5000 GPU.
% You will also need to indicate over how many images, the size and number of dataset, etc.

\subsubsection{Auto-labeling Validation}
Based on the Table \ref{tab:val_acc_comp} and Table \ref{tab:val_acc_comp2}, only slight accuracy difference between the U-Net-Man and U-Net-Auto validates the correctness of our sea ice cover auto-labeling process. 
Figure \ref{fig:validation-auto-man} shows the original S2 image with the manually carried out ground truth label and the prediction generated by the two U-net models, U-Net-Man and U-Net-Auto.

\section{Conclusion}\label{sec:conclusion}
%\section{Conclusions}

%In this research, we explored the color-based segmentation approach to auto-label the Sentinel-2 satellite images for the purpose of providing annotated data for training deep learning-based sea ice classification models. Our results indicate that color-based segmentation provides quite accurate results in labeling thick ice, thin ice, and open water in the polar regions. Therefore, it could be used to auto-label the S2 satellite polar sea ice cover images in the Antarctic in the summer season. 
%We trained two models -  U-Net-Auto on auto-labeled S2 data and U-Net-Man on manually labeled data - for thick ice, thin ice, and open water classification. Both gave high accuracy, and the accuracy difference between the two models was minimal. 
In this work, for 
%thin cloud and shadow-removed color-based 
auto-labeling method, we utilized (i) Python multiprocessing and (ii) PySpark map-reduce-based distributed auto-labeling process on the GCD cluster's heterogeneous systems. Both have achieved almost linear speedups of 4.5x on four physical core machine with hyperthreading, 16.25x on four (one master and three worker) nodes with four cores each respectively, and point to a potential for excellent scalability over larger datasets. 
The distributed U-Net model training on 8 GPUs using Horovod also resulted in a good close-to-linear speedup of 7.21x along with 98.97\% classification accuracy. 
%However, communication overheads within this framework over multi-node GPU cluster currently limits scalability. 

We intend to scale and extend the auto-labeling of larger datasets for various seasons (summer as well as partial summer/winter) and regions (Antarctic and Arctic). We also intend to explore further scalability for distributed model training as well as inferencing over very large datasets on heterogeneous clusters.

%The proposed U-net model inferencing process will be implemented on a Dataproc cluster provided by the Google Cloud service. The cluster configuration will consist of one master node and four worker nodes. In Google Cloud's Dataproc, each machine is equipped with Intel N2 Cascade Lake vCPUs comprising 4 cores. The Spark data frame, UDFs (User Defined Functions), and Spark RDD (Resilient Distributed Dataset) were employed in our study. In order to ensure the functionality of this system over numerous machines, it is necessary to distribute the model among the executor nodes.

%\paragraph{Future Work:} The color range limits for the color-segmentation are not independent of different regions and seasons. For example, we need different brightness limit control for the partial night season as opposed to the summer season reported here. We intend to extend our color-based segmentation and auto-labeling of corresponding datasets for various seasons and regions.
%The U-Net model, on the other hand, is more generic, and we expect to be able to train the model with labeled data over different regions and seasons to achieve a robust sea-ice classification tool independent of seasons and regions.
%We also intend to improve our thin cloud and shadow filter to handle thicker clouds and shadows.

\bibliographystyle{ieeetr}

\bibliography{s2_pdsec_24}

% \bibitem{b1} G. Eason, B. Noble, and I. N. Sneddon, ``On certain integrals of Lipschitz-Hankel type involving products of Bessel functions,'' Phil. Trans. Roy. Soc. London, vol. A247, pp. 529--551, April 1955.
% \bibitem{b2} J. Clerk Maxwell, A Treatise on Electricity and Magnetism, 3rd ed., vol. 2. Oxford: Clarendon, 1892, pp.68--73.
% \bibitem{b3} I. S. Jacobs and C. P. Bean, ``Fine particles, thin films and exchange anisotropy,'' in Magnetism, vol. III, G. T. Rado and H. Suhl, Eds. New York: Academic, 1963, pp. 271--350.
% \bibitem{b4} K. Elissa, ``Title of paper if known,'' unpublished.
% \bibitem{b5} R. Nicole, ``Title of paper with only first word capitalized,'' J. Name Stand. Abbrev., in press.
% \bibitem{b6} Y. Yorozu, M. Hirano, K. Oka, and Y. Tagawa, ``Electron spectroscopy studies on magneto-optical media and plastic substrate interface,'' IEEE Transl. J. Magn. Japan, vol. 2, pp. 740--741, August 1987 [Digests 9th Annual Conf. Magnetics Japan, p. 301, 1982].
% \bibitem{b7} M. Young, The Technical Writer's Handbook. Mill Valley, CA: University Science, 1989.
% \end{thebibliography}

% \vspace{12pt}
% \color{red}
% IEEE conference templates contain guidance text for composing and formatting conference papers. Please ensure that all template text is removed from your conference paper prior to submission to the conference. Failure to remove the template text from your paper may result in your paper not being published.

\end{document}